\documentclass[11pt]{article}

\usepackage[final]{acl}

\usepackage{times}
\usepackage{latexsym}
\usepackage{float}
\usepackage[T1]{fontenc}

\usepackage[utf8]{inputenc}

\usepackage{microtype}

\usepackage{inconsolata}

\usepackage{graphicx}

\usepackage{amsmath,amssymb}
\usepackage{amsthm}
\usepackage{booktabs}
\usepackage{multirow}
\usepackage{xcolor}
\usepackage{enumitem}
\usepackage{subcaption}

\theoremstyle{definition}

\usepackage[most]{tcolorbox}
\tcolorboxenvironment{definition}{
  boxrule=0.5pt,
  colback=gray!5,
  colframe=black!55,
  boxsep=4pt,
  left=6pt, right=6pt, top=6pt, bottom=6pt,
  sharp corners,
  before skip=6pt, after skip=6pt,
}

\newcommand{\cmark}{\textcolor{green!60!black}{\checkmark}}
\newcommand{\xmark}{\textcolor{red}{$\times$}}

\title{InsightEmb: Learning Action-Intent Embeddings\\for Agentic Insight Retrieval}

\author{
    Tsz Ting Chung\textsuperscript{1} \quad
    Jiangnan Li\textsuperscript{2} \quad
    Jie Zhou\textsuperscript{2} \quad
    Mo Yu\textsuperscript{2,$\dagger$}
    \\
    \textsuperscript{1}The Hong Kong University of Science and Technology \\
    \textsuperscript{2}WeChat AI, Tencent 
    }

\begin{document}
\maketitle

\begin{abstract}
Self-improving agents accumulate reusable insights from prior trajectories, making retrieval increasingly important for turning accumulated experience into actionable guidance. 
At each decision step, retrieving the right insight can help the agent progress toward its goal, a setting we refer to as \textbf{agentic insight retrieval}.
However, existing retrieval methods primarily model semantic similarity, while overlooking whether a retrieved insight resolves the agent's current decision bottleneck.
We propose \textbf{InsightEmb}, a contrastive embedding framework that learns transferable progress-oriented retrieval geometry using \emph{only} mathematical reasoning data.
InsightEmb jointly learns to align concrete situations with abstract heuristic rules and to cluster reasoning trajectories with similar progress structures.
We evaluate InsightEmb on dynamic agent tasks and a static skill-retrieval benchmark.
Without any environment-specific training, InsightEmb improves over all these evaluations, surpassing the performance of existing reasoning embedding models.
These results suggest that the geometry of state-insight matching can transfer across domains, enabling effective training from publicly available reasoning data without expensive environment-specific supervision.

\end{abstract}

\section{Introduction}
\label{sec:intro}

LLM-based agents in interactive environments (for example, web navigation~\citep{yao2023webshop}, embodied tasks~\citep{shridhar2021alfworld}, and tool use~\citep{schick2023toolformer}) interleave reasoning and action through frameworks such as ReAct~\citep{yao2023react}.
They must select actions under large action spaces where the optimal strategy depends on abstract reasoning rather than surface-level pattern matching.
Providing agents with \emph{insights}, abstract rules distilled from past experience (e.g., ``check likely locations before exploring randomly''), has proven effective~\citep{majumder2023clin,wang2024voyager,zhao2024expel}, but the agent must dynamically retrieve the most pertinent insight at each step.
We focus on this under-specified retrieval problem: the relevant insight is the one that is \emph{operationally useful now}, given the agent's current state, goal, and action history.

\begin{figure}[t]
\centering
\includegraphics[width=0.48\textwidth]{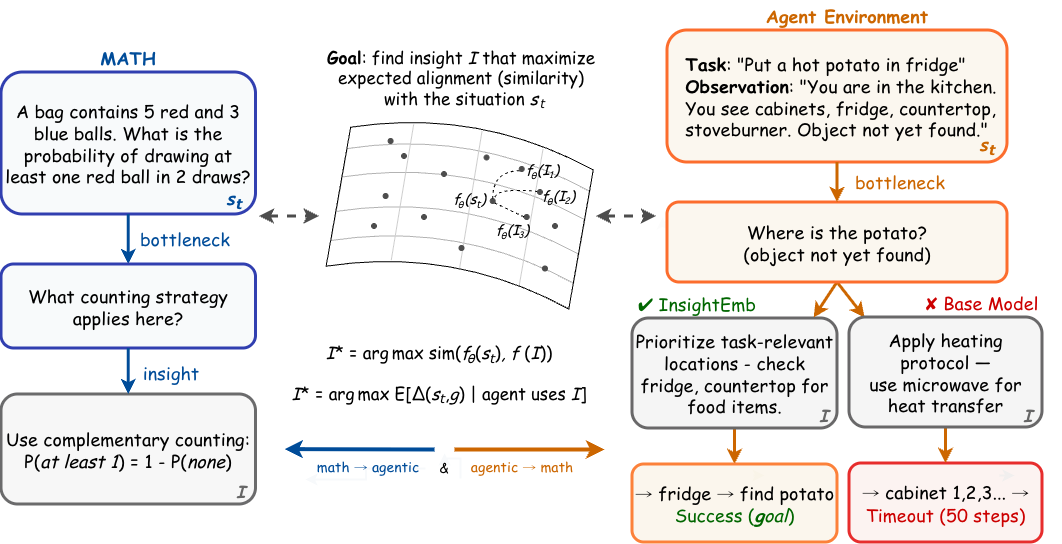 }
\caption{%
InsightEmb retrieves insights that resolve the agent's \emph{current} procedural bottleneck rather than topical overlap alone.
\textbf{Left:} the abstraction gap and shared situation-to-insight structure across math and embodied tasks.
\textbf{Right:} on an ALFWorld ``hot potato'' task, Base favors heating rules before the object is found, whereas InsightEmb retrieves search-first guidance.
}
\label{fig:overview}
\end{figure}

This dynamic retrieval poses a challenge we call the \emph{abstraction gap}: the query (the agent's current observation and action history) and the target (an abstract insight) live at different levels of abstraction, and relevance depends on the agent's \emph{next-step intent} (what bottleneck must be resolved before progress) rather than topical similarity.
Standard embedding models match surface semantics, so off-the-shelf retrievers return topically related but procedurally premature insights, e.g.\ state-transformation rules before the agent has located the target object (Figure~\ref{fig:overview}).

Our key observation is that this \emph{situation-to-insight} matching problem is not domain-specific: both a math query (e.g.\ matching ``$P(\text{at least one red})$'' to complementary counting, with no lexical overlap) and an embodied task (first resolving the \emph{locate-object} bottleneck before heating or placement insights apply) require inferring the latent bottleneck of a concrete situation and retrieving the abstract rule that enables the next effective step.

We propose \textbf{InsightEmb}, a contrastive training framework that exploits this domain-agnostic structure.
InsightEmb trains an embedding model in two stages, \emph{entirely on publicly available mathematical reasoning data}:
(1)~\emph{Situation-to-insight matching} teaches the model to bridge the abstraction gap by mapping math problems and (partial) chain-of-thought trajectories to their relevant heuristic rules.
(2)~\emph{Situation-to-experience matching} teaches structural similarity recognition between reasoning trajectories, reinforcing the model's ability to identify when different-looking situations require the same underlying strategy.
At inference time, the trained model retrieves insights for LLM agents in ALFWorld~\citep{shridhar2021alfworld}, WebShop~\citep{yao2023webshop}, and ScienceWorld~\citep{wang2022scienceworld}, and we use SRA-Bench as a static skill-retrieval diagnostic.
These evaluations test dynamic state-conditioned retrieval and static skill applicability without any environment-specific training data. Our contributions are threefold:
\begin{itemize}[leftmargin=*,nosep]
    \item We formulate agentic insight retrieval as goal-conditioned action-intent matching. Relevance is defined by whether an insight resolves the current bottleneck and enables progress, rather than by semantic similarity alone.
    \item We introduce InsightEmb, a cross-domain contrastive training framework for action-oriented retrieval geometry with empirical validation. 
    \item Our empirical results across static and dynamic evaluations show the effectiveness of InsightEmb with fewer steps taken. Analyses reinforce the gains come from action-intended retrieval.
\end{itemize}

\section{Related Work}
\label{sec:related}

\paragraph{Retrieval-Augmented LLM Agents.}
RAG~\citep{lewis2020rag,guu2020realm} has been extended to agents that retrieve past experiences~\citep{shinn2023reflexion,majumder2023clin,packer2023memgpt}, tool docs~\citep{qin2023toolllm}, and skills~\citep{wang2024voyager}.
These systems target \emph{concrete} artifacts where surface similarity often suffices, whereas we retrieve \emph{abstract} heuristic rules with little lexical overlap, where standard retrievers fail and off-the-shelf or in-domain embedders miss the agentic objective.
A complementary line manages the retrieved context itself, e.g.\ prompt compression~\citep{selectionp2025} and long-context activation approximation~\citep{miasignature2025}, which is orthogonal to \emph{which} insight to retrieve.

\paragraph{Experience-Based Agent Learning.}
Reflexion~\citep{shinn2023reflexion}, CLIN~\citep{majumder2023clin}, and Voyager~\citep{wang2024voyager} accumulate episodic memory, causal abstractions, or skills, while ExpeL~\citep{zhao2024expel} and AutoGuide~\citep{fu2024autoguide} distill trajectories into text rules without weight updates.
They focus on \emph{what} to store, whereas we address \emph{when} to retrieve the right stored knowledge and are agnostic to how insights are produced.
The value of surfacing the right prior experience is further underscored by in-context learning at scale, where retrieved demonstrations drive learning gains~\citep{manyshotcoticl2025}.

\paragraph{Reasoning-Oriented Dense Retrieval.}
Dense retrievers~\citep{karpukhin2020dpr,xiong2021ance,wang2022text,su2023instructor,xiao2023bge} and benchmarks such as MTEB~\citep{muennighoff2022mteb} and BRIGHT~\citep{bright} show that strong semantic matching still leaves reasoning-intensive retrieval far from solved.
ReasonIR~\citep{reasonir}, Llama-NV-Embed-Reasoning~\citep{llama_nv_reasoning} (top open-source \emph{single-model} embedder on BRIGHT), and ReasonEmbed~\citep{bgereasoner} train contrastive embedders for static query--passage support.
Other work makes retrieval \emph{context-aware} through embedding models~\citep{sitemb2025,sitembv15} and memory-aware reranking~\citep{mindscape2025,qfmreranker2025}, but relevance there is still defined by matching a fixed query, not by whether a retrieved item advances an agent toward its goal.
We instead target \emph{next-step utility} under an evolving agent state: whether an insight resolves the current bottleneck, not merely whether it entails a fixed query.

\paragraph{Agentic Benchmarks.}
ALFWorld~\citep{shridhar2021alfworld}, WebShop~\citep{yao2023webshop}, ScienceWorld~\citep{wang2022scienceworld}, Mind2Web~\citep{deng2023mind2web}, and WebArena~\citep{zhou2024webarena} evaluate embodied, scientific, and web control with state-conditioned queries, and AgentBench~\citep{liu2023agentbench} spans additional interactive environments.
Prior evaluation efforts largely fall into two separate camps: retrieval over long contexts~\citep{prelude2025} and reasoning or understanding benchmarks~\citep{stochasticparrot2025,divlogiceval2025,triplethalluc2025}.
A benchmark that jointly targets \emph{retrieval for reasoning and action guidance} is largely missing, BRIGHT~\citep{bright} and SRA-Bench~\citep{su2026skillretrievalaugmentationagentic} begin to address this gap, the latter testing static task-to-skill retrieval over mixed gold and distractor skills and complementing dynamic agent benchmarks by isolating retrieval from online execution.

\section{Method}
\label{sec:method}

\subsection{Problem Setting}

An LLM agent operates in an interactive environment where, at each step~$t$, it observes a state $s_t$ (task description, action history, current observation) and must select an action~$a_t$ toward a goal~$g$, following a reasoning-and-acting loop~\citep{yao2023react}.
The agent has access to an insight corpus $\mathcal{I} = \{I_1, \ldots, I_N\}$, where each insight~$I_i$ is a natural-language description of abstract rules and strategies.
We view this task as \emph{goal-conditioned abductive retrieval}: given the current state and goal, retrieve the insight that best identifies which intervention would make progress possible.
Our notion of relevance is \emph{progress-oriented}: an insight is useful if it is expected to reduce the gap to the goal, not merely if it is semantically similar to the query.
We express this target as,
\begin{equation}
\label{eq:utility_retrieval}
    I^* = \arg\max_{I \in \mathcal{I}} \; \mathbb{E}\left[\Delta(s_t, g) \mid \text{agent uses } I\right],
\end{equation}
where $\Delta(s_t,g)$ measures progress toward goal~$g$ from state~$s_t$.
Equation~\eqref{eq:utility_retrieval} is \emph{not} the training objective, and we do not estimate $\mathbb{E}[\Delta \mid I]$ inside the contrastive loss.
Instead, it guides \emph{how we curate supervision}: for each training problem we run five rollouts, distill candidate insights from random trajectory subsets, and assign an insight to $I^+$ for problem~$q$ only if prepending it to $q$'s query improves solve rate over a no-insight baseline on that same problem (and to $I^-$ otherwise). See Appendix~\ref{app:insight_pipeline}.
At inference, we deploy an embedding retriever that approximates the same criterion by similarity:
\begin{equation}
\label{eq:similarity_retrieval}
    I^* = \arg\max_{I \in \mathcal{I}} \; \text{sim}\bigl(f_\theta(s_t),\; f_\theta(I)\bigr),
\end{equation}
with $f_\theta$ trained by InfoNCE on the curated $(q, I^+, I^-)$ pairs (\S\ref{sec:stage1}).

The core difficulty is the \emph{abstraction gap}: $s_t$ contains concrete details (specific objects, locations, actions) while $I^*$ contains abstract rules (general strategies, heuristics).
For agentic tasks, this gap is also temporal: the relevant insight is the one that addresses the agent's current bottleneck, not necessarily the one most topically related to the final task goal.
For example, before the target object is found, a search insight is more useful than an insight about how to transform or place that object.

We thus define \emph{action-intent} not as a supervised action label but as a progress-conditioned relevance signal tied to Equation~\eqref{eq:utility_retrieval}:

\paragraph{Action-intent embedding}
\label{def:action_intent}
For a situation $s=(g,h,o)$ (goal $g$, action history $h$, observation $o$), the \emph{action-intent} of $s$ is the latent bottleneck that must be resolved before progress toward $g$. An insight $I$ is \emph{action-intent--relevant} to $s$ if it raises expected progress $\mathbb{E}[\Delta(s,g)\mid I]$, and an \emph{action-intent embedding} $f_\theta$ ranks insights by $\mathrm{sim}(f_\theta(s), f_\theta(I))$ under this progress-conditioned relevance rather than topical similarity.

We address this gap through a two-stage contrastive curriculum trained entirely on mathematical reasoning data~\citep{hendrycks2021math,wei2022chain}.
Our training design rests on a single claim: math heuristic retrieval and agentic insight retrieval instantiate the same action-intent matching problem as illustrated in Table~\ref{tab:structural_analogy}.

\subsection{Structural Analogy for Action-Intent Retrieval}
\label{sec:design_claim}

We propose that an agentic action-intent embedding can be learned from math-only data because both settings share the same retrieval structure: a concrete situation must be matched to the abstract rule that resolves the \emph{current} bottleneck, not merely to topically related text.

\begin{table}[t]
\centering
\small
\resizebox{\linewidth}{!}{%
\begin{tabular}{lp{2.4cm}p{2.4cm}}
\toprule
\textbf{Component} & \textbf{Math} & \textbf{Agentic task} \\ \midrule
\textbf{Situation} & Problem statement & Task + observation \\ \midrule
\textbf{Trajectory} & Solution steps & Action history \\ \midrule
\textbf{Insight} & Solving heuristic & Execution strategy \\ \midrule
\textbf{Abstraction gap} & ``at least one red ball'' $\leftrightarrow$ ``complementary counting'' & ``You clean the butterknife'' $\leftrightarrow$ ``place'' \\
\bottomrule
\end{tabular}
}
\caption{Structural analogy between math reasoning and agentic execution, our design claim for building action-intent embeddings from math-only contrastive training.}
\label{tab:structural_analogy}
\end{table}

Under this parallel, contrastive training on math should teach a domain-agnostic geometric property: embed a situation near the abstract rule that makes the next useful step apparent.
The instruction prefix at inference (\S\ref{sec:inference}) frames queries in domain-neutral terms so this property can activate on the target agentic environments without environment-specific fine-tuning.
The experiments below test whether this claim holds in practice.

\paragraph{Shared bottleneck categories.}
What transfers is not a shared ``step'' format but a \emph{situation $\rightarrow$ bottleneck-resolving-rule} geometry organized by a small set of recurring bottleneck \emph{types} that both domains instantiate.
These categories are never specified in the pipeline: the distillation prompts (Appendix~\ref{app:insight_prompts}) ask only for general, task-agnostic rules and never mention any category scheme.
On inspecting the generated math and agentic insights, we find their content maps cleanly onto the same recurring bottleneck types, an emergent property of the distilled insights rather than an artifact engineered into the pipeline.
Reading such a correspondence left-to-right shows the same operation: at a mid-progress state, retrieve the rule that resolves the current bottleneck.
For instance, in the ALFWorld ``put a hot potato in fridge'' case, Base retrieves a topically related but premature heating rule while InsightEmb retrieves the search-priority rule and succeeds, the exact analogue of retrieving ``convert $a \mid b$ into $b \equiv 0 \pmod{a}$'' to resolve the current bottleneck.
Appendix Table~\ref{tab:bottleneck_categories} lists the six emergent bottleneck categories with paired math and agentic examples.

\subsection{Stage 1: Situation-to-Insight Matching}
\label{sec:stage1}

Stage~1 teaches the embedding model to bridge the abstraction gap by matching mathematical problems to their relevant heuristic rules.

\paragraph{Training data.}
Each training example is a triplet $(q, I^+, I^-)$ where $q$ is a query in one of three forms: (i)~a raw math problem statement (\emph{query-only}), (ii)~a problem concatenated with its full chain-of-thought solution (\emph{full trajectory}), or (iii)~a problem with a truncated solution (\emph{partial trajectory}).
The positive set $I^+$ contains distilled heuristic rules that pass the utility filter implied by Equation~\eqref{eq:utility_retrieval}: for each problem in a trajectory subset, a rule is in $I^+$ only if prepending it to that problem's query improved solve rate on validation attempts for that same problem, compared with attempts without the rule.
Rules that fail this check, or hurt performance, are placed in $I^-$.
This ties contrastive labels to \emph{demonstrated progress} rather than author judgment or lexical overlap alone.
For a given trajectory subset, the same $I^+$ and $I^-$ are reused across all three query forms, and only the situation anchor $q$ changes.
Insights are provided at two granularities used jointly in training: \textsc{Bundle} insights (multi-rule summaries distilled from successful and contrasting trajectories) and \textsc{Atomic} insights (single rules split from those bundles).
Stage~1 therefore learns from both coarse strategy sets and fine-grained rules, totaling 11{,}950 contrastive pairs over counting \& probability, number theory, and geometry problems from the MATH dataset~\citep{hendrycks2021math}. Appendix~\ref{app:data_stats} lists pair counts and domain breakdowns.

\paragraph{Training objective.}
Given curated $(q, I^+, I^-)$, we train $f_\theta$ with InfoNCE~\citep{chen2020simclr} and in-batch negatives:
\begin{equation}
\label{eq:infonce}
\small
    \mathcal{L}_1 = -\log \frac{\exp\bigl(\text{sim}(f_\theta(q), f_\theta(I^+)) / \tau\bigr)}{\sum_{I \in \{I^+\} \cup \mathcal{N}} \exp\bigl(\text{sim}(f_\theta(q), f_\theta(I)) / \tau\bigr)}
\end{equation}
where $\tau = 0.01$ and $\mathcal{N}$ includes in-batch negatives (training group size~11).
The loss implements Equation~\eqref{eq:similarity_retrieval} on pairs whose positives were chosen by the progress criterion in Equation~\eqref{eq:utility_retrieval}, not by optimizing $\mathbb{E}[\Delta \mid I]$ end-to-end.

\subsection{Stage 2: Situation-to-Experience Matching}
\label{sec:stage2}

Both stages share the \emph{same} contrastive objective (Equation~\eqref{eq:infonce}) and the same math data, differing only in \emph{what the situation is matched against}, which fixes the level of abstraction the model must bridge:
\begin{itemize}[leftmargin=*,nosep]
\item \textbf{Stage~1 (situation~$\rightarrow$~abstract rule)} matches a situation \emph{across} abstraction levels, to a distilled heuristic that names the bottleneck to resolve.
\item \textbf{Stage~2 (situation~$\rightarrow$~concrete experience)} matches a situation \emph{at} its own abstraction level, to a structurally similar solved problem, teaching the model to recognize when two different-looking situations require the same reasoning approach.
\end{itemize}
The two targets are complementary rather than redundant: Stage~1 supplies the concrete-to-abstract mapping needed to retrieve rules, while Stage~2 sharpens the situation representation itself so that structurally analogous states cluster together, which stabilizes the abstract matching learned in Stage~1.

Concretely, Stage~2 uses triplets $(q, T^+, T^-)$ where $q$ is a raw problem or a full trajectory, $T^+$ is a structurally similar solved problem, and $T^-$ is a dissimilar one, with insights in Stage~1 simply replaced by trajectories.
This stage uses 2{,}896 examples across the same three math domains.

\subsection{Inference: Dynamic Insight Retrieval}
\label{sec:inference}

At inference time, the trained embedding model is deployed for dynamic insight retrieval in target interactive environments \emph{without any domain-specific fine-tuning}.
At each step~$t$, (1)~the agent's current state~$s_t$ is encoded with a task-specific instruction prefix,\footnote{E.g., \emph{``Given an AlfWorld task observation, retrieve relevant insights or strategies that can help the agent solve the task effectively.''}} (2)~the top-$k$ most similar insights are retrieved from the pre-encoded corpus, and (3)~the retrieved insights are prepended to the LLM's prompt for action generation.

\section{Experimental Setup}
\label{sec:experiments}

We fine-tune Qwen3-Embedding-4B~\citep{qwen3embedding} with the two-stage curriculum in \S\ref{sec:method}. Appendix~\ref{app:training_setup} specifies learning rate, batching, contrastive temperature, and hardware, and Table~\ref{tab:hyperparams} summarizes the full configuration.

\subsection{Evaluation Overview}

Our goal is to test the design claim in Table~\ref{tab:structural_analogy}: whether math-only training yields progress-oriented, action-intent retrieval in ALFWorld, WebShop, ScienceWorld, and SRA-Bench.
We use two complementary metrics: \emph{end-task performance} on interactive agent environments and \emph{retrieval recall} on a static benchmark.
On ALFWorld, WebShop, and ScienceWorld, an LLM agent executes full trajectories with dynamic insight retrieval, and we measure whether retrieval improves task completion (success rate and average task score), not ranking accuracy in isolation.
On SRA-Bench, no agent is run. We score task-to-skill matching directly with recall@$k$ and nDCG@$k$.
The three agentic environments (detailed in \S\ref{sec:experiments}) stress complementary procedural skills. Following AgentBench~\citep{liu2023agentbench}, we report success rate for ALFWorld and ScienceWorld and average task score for WebShop.

\subsection{Evaluation with Agentic Environment}

We evaluate dynamic insight retrieval on three text-based interactive environments that stress complementary procedural skills: ALFWorld~\citep{shridhar2021alfworld} (embodied search and manipulation for household tasks), WebShop~\citep{yao2023webshop} (product comparison, variant selection, and multi-step purchase workflows), and ScienceWorld~\citep{wang2022scienceworld} (long-horizon scientific procedures such as measuring melting or boiling points, testing conductivity, and locating living or non-living things).
Each requires a sequence of navigation and interaction actions, making all three direct tests of whether retrieval can select insights that unblock the agent's current procedural phase.

\paragraph{Common agent configuration.}
\label{sec:webshop_setup}\label{sec:scienceworld_setup}
Across all three environments we use the same protocol: Qwen3-8B~\citep{qwen3} as the action-generating LLM with greedy decoding, a history window of 3 steps, a maximum of 50 steps per game, and top-1 dynamic insight retrieval, comparing the same two embedding variants (Base and InsightEmb).
We evaluate ALFWorld on its test split, WebShop on 500 test games, and ScienceWorld on a fixed, seed-42, task-balanced subset of 500 test variations (reused across all experiments). ScienceWorld uses a 700-bundle / 6{,}265-atomic insight pool, and its task-type counts are in Appendix Table~\ref{tab:scienceworld_task_dist}.

\subsubsection{Insight corpora}

We evaluate two insight corpus granularities per environment to disentangle the effect of the embedding model from the insight content.
\textsc{Bundle} insights are multi-rule summaries, each containing several rules with full chain-of-thought reasoning.
\textsc{Atomic} splits each bundle into individual rules with chain-of-thought removed.
Appendix Table~\ref{tab:insight_stats} summarises corpus sizes and rule lengths for each environment.
Appendix~\ref{app:insight_pipeline} documents trajectory collection, DeepSeek-R1/GPT-5.2 distillation prompts, and how \textsc{Bundle} and \textsc{Atomic} corpora are built.

\subsection{SRA-Bench Retrieval Evaluation}
\label{sec:sra_setup}

To isolate retrieval quality from downstream action generation, we additionally evaluate on SRA-Bench~\citep{su2026skillretrievalaugmentationagentic}, spanning theorem proving, logical reasoning, tool use, contest math, medical calculation, and code generation.
Each query uses the full instance \texttt{question} field with a Qwen3-style retrieval instruction prefix~\citep{qwen3embedding}, and each candidate encodes the skill name, description, and body.
Appendix~\ref{app:sra_prompts} gives the exact \texttt{Instruct}/\texttt{Query} template and skill-passage format used in evaluation.
Unlike the agent environments above, the primary metrics here are retrieval recall (R@$k$) and ranking quality (N@$k$) for $k\in\{1,3,5,7,10\}$, macro-averaged across task families (\S\ref{sec:sra_results}).

\section{Results}
\label{sec:results}

\subsection{ALFWorld Results}

We first evaluate whether retrieved insights improve online agent execution in ALFWorld, where the query changes after every observation and action, and on WebShop (\S\ref{sec:webshop_results}), where the agent must search product pages and complete purchases.
Table~\ref{tab:main_results} reports both metrics under the same retrieval protocol.

\begin{table}[t]
\centering
\small
\setlength{\tabcolsep}{4pt}
\begin{tabular}{llccc}
\toprule
\textbf{Insight} & \textbf{Embedding} & \textbf{ALF} & \textbf{WS} & \textbf{SW} \\
\midrule
\{\} & --- & 52.86 & 31.03 & 2.40 \\
\midrule
\textsc{Bundle} & Base & 54.29 & 18.42 & 7.40 \\
\textsc{Bundle} & InsightEmb & \textbf{60.71} & \textbf{31.74} & \textbf{8.00} \\
\midrule
\textsc{Atomic} & Base & 55.00 & 28.25 & 7.40 \\
\textsc{Atomic} & InsightEmb & \textbf{59.29} & \textbf{32.05} & \textbf{10.20} \\
\bottomrule
\end{tabular}
\caption{Dynamic agent evaluation on ALFWorld (ALF), WebShop (WS), and ScienceWorld (SW) (top-1 retrieval, Qwen3-8B agent). All columns are reported as percentages (success rate for ALF and SW, average task score for WS). The no-insight baseline for SW is only $2.40$, so every insight setting more than triples it. Separate SW top-$k$ curves appear in Appendix Table~\ref{tab:scienceworld_topk}.}
\label{tab:main_results}
\end{table}

\paragraph{InsightEmb transfers to embodied control.}
All InsightEmb variants outperform the Base embedding model on test success rate in Table~\ref{tab:main_results}.
With both \textsc{Bundle} and \textsc{Atomic} insights, InsightEmb achieves about a 5-point improvement over Base.
For \textsc{Bundle}, per-game-level retrieval turnover, step counts, and zero-score rates are in Appendix Table~\ref{tab:episode_stats} (\S\ref{app:episode_alf}).
Since InsightEmb is trained exclusively on mathematical reasoning data, these gains are entirely from cross-domain transfer.
Comparison with an ALFWorld-trained in-domain retriever is deferred to \S\ref{sec:indomain}.

\paragraph{Per-task-type breakdown.}
Appendix Table~\ref{tab:per_task} (within \S\ref{app:alfworld_analysis}) breaks down test success rates by ALFWorld task type.
InsightEmb leads on 4 of 6 types, with the largest gains on \emph{clean} and \emph{find\_two}, indicating a general structural-matching gain rather than a single-task shortcut.

\paragraph{Mechanistic analysis.}
Appendix~\S\ref{app:episode_stats}--\S\ref{app:step_dynamics} analyzes why retrieval helps on ALFWorld: per-game statistics, qualitative cases, topical vs.\ procedural rule matching, and step-conditioned retrieval dynamics. InsightEmb shows a better retrieval diversity, step counts, and lower zero-score rates, and a state-aware transition from search to verification and state transformation.

\subsection{WebShop Results}
\label{sec:webshop_results}

WebShop results are summarized in Table~\ref{tab:main_results} (right column).
We report average task score as a percentage, which reflects partial credit for attribute matching under the AgentBench evaluation protocol.

\paragraph{InsightEmb transfers to web shopping.}
On WebShop, retrieval quality matters: \textsc{Bundle} with Base (18.42\%) falls well below the no-insight baseline (31.03\%), while InsightEmb achieves a better performance (\textsc{Bundle} 31.74\%, \textsc{Atomic} 32.05\%).
InsightEmb outperforms Base on both corpora (e.g., +72\% relative on \textsc{Bundle}). Per-game-level statistics for \textsc{Bundle} are in Appendix Table~\ref{tab:episode_stats} (\S\ref{app:episode_webshop}).

\paragraph{Retrieval safety under a noisy insight pool.}
The WebShop \textsc{Bundle} regime highlights a distinct property beyond average relevance: a progress-blind retriever can be \emph{worse than no retrieval at all} (Base \textsc{Bundle} 18.42\% vs.\ 31.03\% no-retrieval), because it surfaces premature or topically-related but procedurally irrelevant insights from a largely unhelpful pool.
InsightEmb learns to filter such premature insights and select the state-appropriate one, turning a net-harmful retrieval setting into a net-useful one (31.74\%).
Making agentic insight retrieval \emph{safe} under a noisy corpus is itself a useful property, not only improving average relevance but also avoiding active degradation of the base agent.

\paragraph{Qualitative patterns.}
Manual inspection of divergent games reveals three recurring patterns (variant-selection awareness, loop prevention, and procedural sequencing), indicating that InsightEmb performs \emph{procedural matching}, retrieving insights that address the agent's current bottleneck. Appendix~\ref{app:webshop_examples} details each pattern with side-by-side Base vs.\ InsightEmb trajectories.

\subsection{ScienceWorld Results}
\label{sec:scienceworld_results}

\paragraph{InsightEmb transfers to scientific procedures.}
ScienceWorld results (top-1 retrieval) are summarized in Table~\ref{tab:main_results} (right column).
This environment is substantially harder for the base agent: the no-insight baseline reaches only $2.40\%$ success, reflecting the long-horizon, multi-step nature of scientific procedures (measurement, classification, and controlled experiments).
Against this backdrop InsightEmb wins or ties Base in both corpora at top-1 ($10.20\%$ vs.\ $7.40\%$ on \textsc{Atomic} and $8.00\%$ vs.\ $7.40\%$ on \textsc{Bundle}), and every insight setting more than triples the no-insight baseline.
Since InsightEmb is trained only on mathematical reasoning data, these gains on a third, procedurally distinct environment are again entirely from cross-domain transfer.

\paragraph{Robustness across budget and action model.}
The advantage persists across retrieval budgets $k\in\{1,3,5\}$ (Appendix Table~\ref{tab:scienceworld_topk}) and under a stronger closed-source action model (GPT-5.2, Appendix Table~\ref{tab:gpt52_cross_env}), mirroring the budget- and model-robustness checks reported for ALFWorld and WebShop.

\paragraph{Per-task-type grouping.}
The ScienceWorld task types span complementary functional families (search/identify, measurement/verification, state-transform/experiment, and multi-step reasoning, detailed in Appendix~\ref{app:scienceworld_per_task_sec}), so the aggregate gain reflects broad procedural coverage rather than a single task shortcut.

\paragraph{Mechanistic analysis.}
Per-game-level statistics (\textsc{Bundle}, top-1) are reported alongside ALFWorld and WebShop in Appendix Table~\ref{tab:episode_stats} (\S\ref{app:episode_scienceworld}).
Aside from the task metric, InsightEmb lowers the negative-score rate from $55.4\%$ to $52.0\%$ and raises mean distinct insights from 2.04 to 2.37.
Notably, on ScienceWorld InsightEmb more often locks onto a single procedurally complete workflow insight and reuses it across an experiment. Its higher mean step count ($40.96 \rightarrow 43.59$) indicates it more often persists through the multi-step protocol rather than terminating early without completing the required state transition.

\paragraph{Qualitative patterns.}
Manual inspection of divergent games reveals three recurring patterns (state-variable awareness, action sequencing, and failure-mode avoidance), mirroring the procedural-matching behavior seen on ALFWorld and WebShop. Appendix~\ref{app:scienceworld_examples} details each pattern with side-by-side Base vs.\ InsightEmb trajectories and worked examples.

\subsection{Static Skill Retrieval on SRA-Bench}
\label{sec:sra_results}

Following the static protocol in \S\ref{sec:sra_setup}, we test whether math-only training improves \emph{task-to-skill} matching before any action model runs.
Table~\ref{tab:sra_macro} reports macro-averaged results over the 636 gold skills and 26{,}262 distractors, and the per-task-family breakdown at @1 and @10 is in Appendix Table~\ref{tab:sra_per_task} (\S\ref{app:sra_analysis}).
InsightEmb improves over Base on every cutoff, with small gains at R@1 and substantially larger gains at higher cutoffs (+8.90 R@10).
This pattern indicates that the training primarily improves the ranking of relevant skills beyond the first position, which is especially useful when an agent can inspect or condition on multiple retrieved skills.

\begin{table*}[t]
\centering
\small
\begin{tabular}{lcccccccccc}
\toprule
\textbf{Embedding} & \textbf{R@1} & \textbf{R@3} & \textbf{R@5} & \textbf{R@7} & \textbf{R@10} & \textbf{N@1} & \textbf{N@3} & \textbf{N@5} & \textbf{N@7} & \textbf{N@10} \\
\midrule
Base & 30.33 & 40.84 & 46.40 & 49.83 & 54.13 & 33.82 & 37.62 & 39.96 & 41.23 & 42.64 \\
InsightEmb & \textbf{31.37} & \textbf{46.39} & \textbf{54.25} & \textbf{58.56} & \textbf{63.03} & \textbf{36.29} & \textbf{42.00} & \textbf{45.29} & \textbf{46.92} & \textbf{48.42} \\
\midrule
$\Delta$ & +1.04 & +5.55 & +7.85 & +8.73 & +8.90 & +2.47 & +4.38 & +5.33 & +5.69 & +5.78 \\
\bottomrule
\end{tabular}
\caption{SRA-Bench macro-average retrieval results across six task families. Queries use task information and candidates use the full skill content. R@$k$ is recall at $k$ and N@$k$ is normalized discounted cumulative gain at $k$. Appendix Table~\ref{tab:sra_per_task} breaks down per-task-family results at @1 and @10, and the macro-average is computed by averaging those task families equally.}
\label{tab:sra_macro}
\end{table*}

Across families, InsightEmb improves clearly on theorem proving, tool use, contest math, and code generation, where queries and useful skills differ in wording while sharing procedural structure. The only clear negative outlier is MedCalcBench at small cutoffs, whose highly template-like, lexically specialized skills favor the Base embedder's entity-level cues at rank~1 (Appendix Table~\ref{tab:sra_per_task}).
Appendix~\ref{app:sra_analysis} discusses the per-family breakdown and the MedCalcBench outlier, lexical vs.\ structural retrieval bottlenecks, and implications for reranker-limited skill pipelines, and Appendix~\ref{app:medcalc_bm25} provides a lightweight BM25-hybrid remedy that recovers R@1 and reaches $100\%$ recall at R@5/R@10 on this family, together with an explicit applicability scope.

\section{Analysis and Comparisons}
\label{sec:analysis}

The results above establish that math-trained InsightEmb improves dynamic agent execution and static skill retrieval over the Base embedder.
This section asks \emph{why}: we compare reasoning-oriented embedders and alternative insight generators, validate in-domain fine-tuning against cross-domain training, then analyze whether the learned geometry matches our structural-analogy claim.

\subsection{Scaling with Comparison to Reasoning-Oriented Retrievers}
\label{sec:reasoning_retrievers}

We focus this analysis on ALFWorld and WebShop because they represent the two opposite \emph{insight-pool regimes} that a retriever must handle, as revealed by comparing the no-insight baseline (\{\}) with Base-selected insights in Table~\ref{tab:main_results}.
On ALFWorld, retrieval is \emph{helpful even with an untrained retriever}: adding Base-selected \textsc{Bundle} insights already raises success over no insight ($52.86 \rightarrow 54.29$), so the corpus is a largely \emph{useful} insight pool and the question is how much better a progress-aware retriever can do.
On WebShop, the same Base-selected insights are \emph{harmful}: \textsc{Bundle}+Base drops well below no insight ($31.03 \rightarrow 18.42$), indicating a corpus that is largely \emph{unhelpful or premature}, where a progress-blind retriever surfaces net-negative insights.
Studying both regimes together therefore tests two distinct demands, extracting more value from a useful pool (ALFWorld) and staying safe under a noisy pool (WebShop), which is why we concentrate the scaling comparison here rather than on a single environment.

\begin{figure}[t]
\centering
\includegraphics[width=\columnwidth]{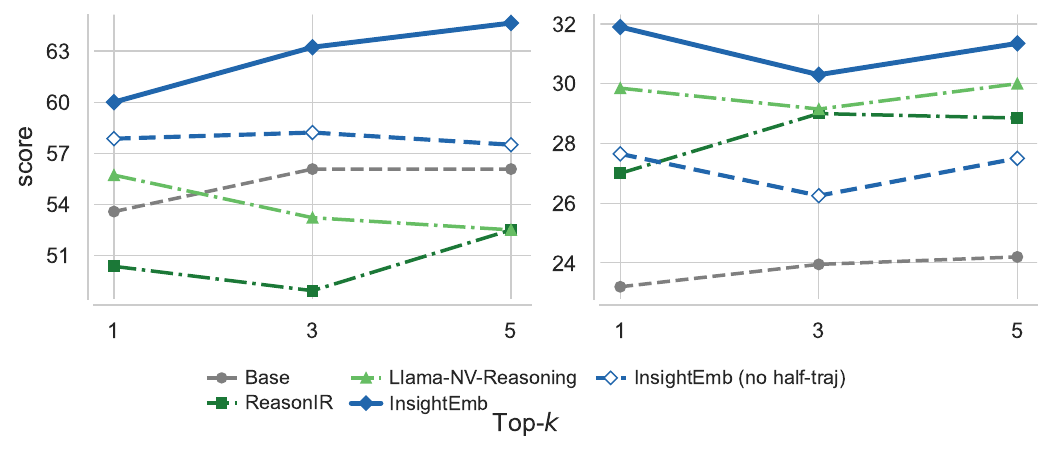}
\caption{Top-$k$ scaling for reasoning-oriented retrievers. \textbf{Left:} ALFWorld success rate . \textbf{Right:} WebShop average task score . Curves show the arithmetic mean of \textsc{Bundle} and \textsc{Atomic} retrieval at each $k \in \{1,3,5\}$ for Base, ReasonIR, Llama-NV-Reasoning, and InsightEmb (solid). The dashed line (InsightEmb color) is the \emph{no half-traj} ablation, which removes partial-trajectory queries from Stage~1.}
\label{fig:reasoning_baselines_corpora_average}
\end{figure}

A natural question is whether reasoning-oriented retrievers trained for static, query--document matching can serve as drop-in replacements for InsightEmb in these two regimes.
Using the same ALFWorld and WebShop agent protocol as Table~\ref{tab:main_results}, we compare Base, InsightEmb, and two reasoning-oriented embedding baselines:
(i)~\textbf{Llama-NV-Embed-Reasoning-3B}~\citep{llama_nv_reasoning} (3.2B, Llama-3.2-3B), the top-ranked open-source single-model embedder on BRIGHT, and
(ii)~\textbf{ReasonIR-8B}~\citep{reasonir} (8B, Llama-3.1-8B), trained with contrastive learning on synthetic reasoning-intensive pairs, showing the setting closest to ours.
Both differ from InsightEmb (4B, Qwen3-Embedding-4B~\citep{qwen3embedding}), which is trained only on math data.
Beyond the top-1 results in Table~\ref{tab:main_results}, we sweep retrieval budgets $k \in \{1, 3, 5\}$ only for all four embedders on both environments and both insight corpora (\textsc{Bundle} and \textsc{Atomic}).
At $k{=}5$, top-\textsc{Bundle} retrieval already prepends about 25 instruction-level rules to the agent prompt (roughly five rules per bundle), so we do not evaluate larger $k$.
Figure~\ref{fig:reasoning_baselines_corpora_average} plots ALFWorld  (\textbf{left}) and WebShop  (\textbf{right}) performances as a function of $k$, using the arithmetic mean of \textsc{Bundle} and \textsc{Atomic} at each $k \in \{1,3,5\}$.
We also ablate Stage~1 with a \emph{no half-traj} variant that drops partial-trajectory queries (dashed curve). It lies between Base and full InsightEmb, accounting for most of the Base-to-InsightEmb gain (Appendix~\ref{app:no_halftraj}). Appendix~\ref{app:reasoning_topk} also provides separate ALFWorld/WebShop top-$k$ curves by corpus.

On the corpus-averaged curves in Figure~\ref{fig:reasoning_baselines_corpora_average}, InsightEmb surpasses ReasonIR and Llama-NV-Reasoning in both ALFWorld and WebShop at each $k \in \{1,3,5\}$.
This gap is not explained by stronger static matching alone: reasoning-oriented retrievers are trained for \emph{static} query--passage matching, whereas agentic insight retrieval requires \emph{dynamic action-intent} matching against an evolving state to retrieve the rule that unblocks the current procedural step, the geometry InsightEmb is trained for and they are not.

\subsection{Robustness to Insight Generator and Action Model}
\label{sec:gpt52_insights}

Our main results use DeepSeek-R1~\citep{deepseekr1} to distill insights and Qwen3-8B as the action model. We verify that the gains are not tied to either choice.
\emph{Alternative insight generator:} repeating the dynamic-agent evaluation with GPT-5.2~\citep{openai2025gpt52}-generated corpora (same action model and top-$k$ protocol), InsightEmb consistently outperforms Base on both ALFWorld and WebShop across $k \in \{1,3,5\}$ (Appendix~\ref{app:gpt52_topk}, Figure~\ref{fig:gpt52_insights_corpora_average}).
\emph{Alternative action model:} replacing Qwen3-8B with the stronger closed-source GPT-5.2 as the action model preserves the advantage across all three environments (Appendix~\ref{app:scienceworld_gpt52}, Table~\ref{tab:gpt52_cross_env}).
Together, the improvement holds across insight source and action model.

\subsection{Validating the Structural-Analogy Claim}
\label{sec:bidirectional}
\label{sec:indomain}

\paragraph{Math-to-agentic transfer.} Table~\ref{tab:structural_analogy} predicts \emph{bidirectional} transfer: math training should improve agentic retrieval (tested in \S\ref{sec:results}), and with this hypothesis, embodied training should improve math retrieval also.
We validate both directions and compare against the natural alternative of in-environment fine-tuning.

\paragraph{Finetuning With ALFWorld} A standard alternative to cross-domain training is to fine-tune the embedder on ALFWorld situation-to-insight pairs (same backbone, pipeline, and top-1 agent stack as \S\ref{sec:experiments}).
As Table~\ref{tab:indomain_alf} shows, InsightEmb exceeds this \emph{In-domain} retriever on both corpora despite using no ALFWorld data, and despite the in-domain set being \emph{larger} (21{,}318 vs.\ 14{,}846 math pairs), so the gap reflects the supervision source, not scale.
Appendix~\ref{app:math_vs_indomain} makes this precise by \emph{measuring} the geometric diversity of the two sources (pairwise/nearest-neighbor cosine distance and covariance effective rank), replacing the informal notion of ``structural diversity''.

\begin{table}[t]
\centering
\small
\begin{tabular}{llc}
\toprule
\textbf{Corpus} & \textbf{Embedding} & \textbf{ALFWorld} \\
\midrule
\multirow{3}{*}{\textsc{Bundle}}
& Base & 54.29 \\
& ALF (In-domain) & \textbf{57.86} \\
& InsightEmb & \textbf{60.71} \\
\midrule
\multirow{3}{*}{\textsc{Atomic}}
& Base & 55.00 \\
& ALF (In-domain) & \textbf{57.86} \\
& InsightEmb & \textbf{59.29} \\
\bottomrule
\end{tabular}
\caption{ALFWorld test success rate  for an ALFWorld-trained in-domain retriever (\textsc{Bundle} only) vs.\ Base and math-trained InsightEmb (top-1 retrieval, Qwen3-8B agent). The in-domain model is trained and evaluated on the same environment.}
\label{tab:indomain_alf}
\end{table}

\paragraph{Agentic-to-math transfer.}
For the reverse direction, we evaluate all model variants on \emph{math insight retrieval}, retrieve the correct heuristic rule for a held-out math problem. Table~\ref{tab:math_insight} presents accuracy on four math domains.
Three domains (number theory, geometry, and counting \& probability) overlap with the training distribution (same domains, different problems from the test split with a separate insight corpus), while algebra is entirely out-of-distribution and never appears in training. InsightEmb consistently improves over Base on the held-out math test set and the unseen math domain. This confirms that the contrastive curriculum does not sacrifice in-domain math retrieval while learning the claimed geometry. More importantly, the ALFWorld-trained model (ALF) also improves over Base on math insight retrieval across all four domains. This reverse-direction gain supports Table~\ref{tab:structural_analogy}: the geometry is shared, not a one-way math-to-agent shortcut.

\begin{table}[t]
\centering
\small
\begin{tabular}{lccc}
\toprule
\textbf{Task} & \textbf{Base} & \textbf{InsightEmb} & \textbf{ALF} \\
\midrule
Number Theory$^*$ & 77.22 & \textbf{79.81} & \textbf{78.33} \\
Geometry$^*$ & 54.07 & \textbf{59.29} & \textbf{57.41} \\
Counting \& Probability$^*$ & 74.47 & \textbf{77.43} & \textbf{75.53} \\
Algebra & 86.69 & \textbf{88.21} & \textbf{87.81} \\
\bottomrule
\end{tabular}
\caption{Math \emph{insight} retrieval accuracy. ALF = ALFWorld-trained model. $^*$In-domain (same domains as training, held-out test split with separate insight pool).}
\label{tab:math_insight}
\end{table}

\section{Conclusion}
\label{sec:conclusion}

We studied \emph{agentic insight retrieval}: at each decision step, an agent must retrieve an abstract rule that resolves its \emph{current} procedural bottleneck, not merely text similar to the observation.
We introduced \textbf{InsightEmb}, a contrastive framework that learns goal-conditioned action-intent matching from mathematical reasoning data alone.
On ALFWorld, WebShop, and ScienceWorld it raises \emph{end-task} success under dynamic top-$k$ insight injection, and on SRA-Bench it raises \emph{retrieval recall} and nDCG for task-to-skill matching, outperforming the Base embedder, an ALFWorld-trained in-domain retriever, and strong reasoning-oriented embedders.
The gains are consistent across environments, action models (Qwen3-8B and GPT-5.2), insight sources (DeepSeek-R1, GPT-5.2, prebuilt skills), corpus granularities, and retrieval budgets. InsightEmb further makes retrieval \emph{safe} under a noisy insight pool where a progress-blind retriever can be worse than no retrieval at all.
Together, these results support a simple design claim: math heuristic retrieval and agentic insight retrieval share the same progress-oriented matching structure.

\paragraph{Limitations.}
Although SRA-Bench directly evaluates static retrieval quality, dynamic agentic retrieval lacks gold labels identifying the optimal insight at each state. We therefore use downstream task success as an indirect but objective-aligned measure of retrieval utility. In addition, the current utility labels for $(I^+, I^-)$ rely on a single validation attempt for each candidate insight--problem pair, so stochasticity in LLM inference may introduce false-positive or false-negative labels. Moreover, although we train with both full and partial trajectories, the partial-trajectory anchors currently reuse problem-level $(I^+, I^-)$ labels rather than state-specific supervision. Nevertheless, the current design already yields substantial and consistent improvements across static and dynamic evaluations. We leave repeated validation and the construction of state-specific training pairs to future work, which may further improve label reliability and state awareness.

\bibliography{custom}

\clearpage
\appendix

\section{Training Data and Insight Corpora}
\label{app:data_corpora}

\subsection{Training Data Statistics}
\label{app:data_stats}

InsightEmb is trained only on mathematical reasoning data (Stage~1: situation-to-insight, Stage~2: situation-to-experience, \S\ref{sec:stage1}--\ref{sec:stage2}).
Table~\ref{tab:data_stats} summarizes contrastive training examples per stage.
Stage~1 combines \textbf{bundle} insight pairs (7{,}740 unique rules in the insight pool) and \textbf{atomic} pairs (22{,}243 unique rules).
``Query'' = query-only samples, and ``Traj'' = query concatenated with a full or partial CoT trajectory.
Table~\ref{tab:task_dist} gives the per-domain Stage~1 breakdown (bundle + atomic).

\begin{table}[H]
\centering
\small
\resizebox{\linewidth}{!}{%
\begin{tabular}{llccc}
\toprule
\textbf{Stage} & \textbf{Granularity} & \textbf{Total} & \textbf{Query} & \textbf{Traj} \\
\midrule
\multirow{3}{*}{Stage 1}
& Bundle & 5{,}463 & 1{,}917 & 3{,}546 \\
& Atomic & 6{,}487 & 2{,}308 & 4{,}179 \\
& \textbf{Combined} & \textbf{11{,}950} & \textbf{4{,}225} & \textbf{7{,}725} \\
\midrule
Stage 2 (CoT) & --- & 2{,}896 & 2{,}329 & 567 \\
\bottomrule
\end{tabular}
}
\caption{Training data statistics. Stage~1 totals combine bundle and atomic insight pairs.}
\label{tab:data_stats}
\end{table}

\begin{table}[H]
\centering
\small
\resizebox{\linewidth}{!}{%
\begin{tabular}{lccc}
\toprule
\textbf{Math Domain} & \textbf{Bundle} & \textbf{Atomic} & \textbf{Stage 1 Total} \\
\midrule
Counting \& Probability & 1{,}736 & 2{,}647 & 4{,}383 \\
Number Theory & 2{,}220 & 2{,}400 & 4{,}620 \\
Geometry & 1{,}507 & 1{,}440 & 2{,}947 \\
\midrule
\textbf{Total} & 5{,}463 & 6{,}487 & 11{,}950 \\
\midrule
\textbf{Math Domain} & \multicolumn{3}{c}{\textbf{Stage 2 (CoT)}} \\
\midrule
Counting \& Probability & \multicolumn{3}{c}{966} \\
Number Theory & \multicolumn{3}{c}{966} \\
Geometry & \multicolumn{3}{c}{964} \\
\midrule
\textbf{Total} & \multicolumn{3}{c}{2{,}896} \\
\bottomrule
\end{tabular}
}
\caption{Per-domain training counts. Stage~1 = bundle + atomic insight pairs, Stage~2 = CoT experience pairs.}
\label{tab:task_dist}
\end{table}

\subsection{Insight Corpus Construction}
\label{app:insight_pipeline}

This appendix documents how agentic and math insight corpora are built and how they are converted into InsightEmb training data.

\subsubsection{Pipeline Overview}

\textbf{Step 1: Collect training trajectories.}
For each math training problem (or ALFWorld/WebShop training game), we run the task model five times on the training split and log full interaction histories, following trajectory-collection practice in self-improvement methods~\citep{zelikman2022star,zhao2024expel}.
Successful and failed rollouts are grouped by problem, retaining all five responses per instance together with labels indicating which rollouts succeeded.

\textbf{Step 2: Distill insights.}
We apply a multi-subset genetic search over trajectory subsets.
For each subset, an LLM (DeepSeek-R1~\citep{deepseekr1} for main experiments, GPT-5.2~\citep{openai2025gpt52} for the robustness check) receives a distillation prompt (\S\ref{app:insight_prompts}) and produces a \textsc{Bundle} insight: a numbered list of general, task-agnostic rules.
Each candidate insight is validated on the problems represented in that subset: we prepend the rules to each problem's query, re-run the solver, and keep the insight only if it improves solve rate over a no-insight baseline on those problems.
Chromosomes that yield helpful bundles are kept, and \textsc{Atomic} rules are obtained by splitting each bundle into individual numbered rules (chain-of-thought removed).
For ALFWorld and WebShop, each trajectory step is first compressed into a one-sentence summary before distillation.

\textbf{Step 3: Build Stage~1 embedding training pairs.}
Validated bundle and atomic insights are paired with math problems in two query forms: query-only (problem statement) and query+trajectory (problem with full or partial chain-of-thought).
Each pair assigns $I^+$ to insights that improved solve rate on that problem in the subset-validation check above, and $I^-$ to insights that did not (Equation~\eqref{eq:utility_retrieval}), then filters pairs to ensure enough contrastive negatives per training group.

\textbf{Step 4: Build Stage~2 CoT training pairs.}
We construct situation-to-experience pairs from structurally similar solved problems, pairing each query with helpful and unhelpful demonstration trajectories.

\subsubsection{Insight-Generation Prompts}
\label{app:insight_prompts}

\paragraph{Math (successful trials).}
Used when all trajectories in a subset succeeded:
\begin{quote}\small
You are an advanced reasoning agent that can create rules based on forming new critiques of past task trajectories.

You will be given successful tasks trials in which you are doing math derivation.
Here are the trails.
\{trajs\}
By examining the successful trials, you can create the new rules are GENERAL and HIGH LEVEL insights of the successful trials or proposed way of Thought so they can be used as helpful tips to different tasks in the future.
Have an emphasis on tips that help the agent perform better Thought and Action.
Do not mention the trials in the rules because all the rules should be GENERALLY APPLICABLE.
Each rule should be concise and easy to follow.
Any operation can be used MULTIPLE times.
\end{quote}

\paragraph{Math (mixed success and failure).}
Used when a subset contains both correct and incorrect solutions, with trajectories wrapped as \verb|<correct_trial_i>| or \verb|<wrong_trial_i>|:
\begin{quote}\small
You are an advanced reasoning agent that can create rules based on forming new critiques of past task trajectories.

You will be given multiple tasks trials in which you are doing math derivation.
Here are the trails.
\{trajs\}
By examining and contrasting between failed and successful trials, you can create the new rules are GENERAL and HIGH LEVEL critiques of the failed trials or proposed way of Thought so they can be used as helpful tips or to avoid similar failures when encountered with different questions in the future.
Have an emphasis on critiquing how to perform better Thought and Action.
Do not mention the trials in the rules because all the rules should be GENERALLY APPLICABLE.
Each rule should be concise and easy to follow.
Any operation can be used MULTIPLE times.
\end{quote}

\paragraph{ALFWorld (successful trials).}
After step-level summarization, used when all trajectories in a subset succeeded:
\begin{quote}\small
You are an advanced reasoning agent that can create rules based on forming new critiques of past AlfWorld task trajectories.

You will be given successful AlfWorld task trials involving multi-turn conversations with an environment.
Here are the trajectories:
\{trajs\}

By examining the successful trials, you can create new rules that are GENERAL and HIGH LEVEL insights about successful AlfWorld task completion strategies.
Focus on insights that help the agent perform better in multi-turn dialogue with environments.

Key aspects to consider:
- Effective communication strategies with the environment
- Problem-solving approaches in interactive settings
- Handling of multi-step reasoning in dialogue
- Adaptation to environment feedback
- Efficient exploration and action selection

Do not mention specific trials in the rules because all rules should be GENERALLY APPLICABLE to AlfWorld tasks.
Each rule should be concise and easy to follow.
Any operation can be used MULTIPLE times.
\end{quote}

\paragraph{ALFWorld (mixed success and failure).}
\begin{quote}\small
You are an advanced reasoning agent that can create rules based on forming new critiques of past AlfWorld task trajectories.

You will be given multiple AlfWorld task trials involving multi-turn conversations.
Here are the trajectories:
\{trajs\}

By examining and contrasting between failed and successful trials, you can create new rules that are GENERAL and HIGH LEVEL critiques of failed strategies or proposed ways of thinking.
Focus on critiquing how to perform better in multi-turn dialogue with environments.

Key aspects to consider:
- Communication breakdowns with the environment
- Ineffective problem-solving approaches
- Poor handling of multi-step reasoning
- Failure to adapt to environment feedback
- Inefficient exploration strategies

Do not mention specific trials in the rules because all rules should be GENERALLY APPLICABLE.
Each rule should be concise and easy to follow.
\end{quote}

\paragraph{WebShop (successful trials).}
After step-level summarization, used when all trajectories in a subset succeeded:
\begin{quote}\small
You are an advanced reasoning agent that can create rules based on forming new critiques of past WebShop task trajectories.

You will be given successful WebShop task trials involving multi-turn conversations with a web shopping environment.
Here are the trajectories:
\{trajs\}

By examining the successful trials, you can create new rules that are HIGH LEVEL insights about successful WebShop task completion strategies.
Focus on insights that help the agent perform better in multi-turn dialogue with web shopping environments.

Key aspects to consider:
- Effective search query formulation strategies
- Product selection and comparison approaches
- Handling of product attributes (size, color, price, etc.)
- Navigation strategies (search, click, back, buy)
- Efficient exploration of product listings
- Matching user requirements to product descriptions
- Decision-making on when to buy vs.\ continue searching

Do not mention specific trials in the rules.
Each rule should be concise and easy to follow.
Any operation can be used MULTIPLE times.
\end{quote}

\paragraph{WebShop (mixed success and failure).}
\begin{quote}\small
You are an advanced reasoning agent that can create rules based on forming new critiques of past WebShop task trajectories.

You will be given multiple WebShop task trials involving multi-turn conversations.
Here are the trajectories:
\{trajs\}

By examining and contrasting between failed and successful trials, you can create new rules that are HIGH LEVEL critiques of failed strategies or proposed ways of thinking.
Focus on critiquing how to perform better in multi-turn dialogue with web shopping environments.

Key aspects to consider:
- Ineffective search query formulation
- Poor product selection and comparison decisions
- Failure to match product attributes to requirements
- Inefficient navigation patterns
- Premature or delayed purchase decisions
- Misinterpretation of product descriptions
- Failure to adapt search strategy based on results

Do not mention specific trials in the rules.
Each rule should be concise and easy to follow.
\end{quote}

\paragraph{Insight validation (math).}
For a subset of trajectories, the distillation prompt produces a candidate rule set, which we then test on every problem in that subset by prepending the rules to the problem query and re-running the solver:
\begin{quote}\small
Below is an instruction that describes a task.
Write a response that appropriately completes the request and wrap the final answer inside \textbackslash boxed\{\{\}\}.

\{generated insight rules\}
\#\#\# Problem:
\{problem from the subset\}

\#\#\# Solution: Let's think step by step.
\end{quote}
For each problem~$q$ in the subset, we record whether the solver succeeds with vs.\ without the prepended rules. The insight is assigned to $I^+$ on problems where it helps and to $I^-$ on problems where it does not, which defines the Stage~1 contrastive labels.

\subsubsection{Example Generated Insights}
\label{app:insight_examples}

\paragraph{Math (counting \& probability).}
\begin{quote}\small
- \textbf{Define Variables \& Relationships}: Start by defining variables for unknown quantities and establish mathematical relationships based on given conditions to structure the problem.

- \textbf{Complementary Probability for ``Not'' Events}: When calculating the probability of an event \textbf{not} occurring (e.g., not sitting together), compute the probability of the complementary event and subtract it from 1.

- \textbf{Leverage Standard Combinatorial Formulas}: Recognize scenarios like distributing identical items (stars and bars) or non-colinear points (triangles in a cube) and apply the appropriate formulas directly.
\end{quote}

\paragraph{ALFWorld.}
\begin{quote}\small
1.\ \textbf{Maintain a Goal-Driven Action Sequence}: Successful agents follow a clear path toward the goal, while failed ones get sidetracked.

2.\ \textbf{Adapt Exploration Based on Feedback}: If an object isn't found where expected, systematically check other plausible locations instead of repeating the same action.

3.\ \textbf{Verify Object Presence Before Interaction}: Avoid assuming an object is present, and check first to prevent unnecessary actions.

4.\ \textbf{Prioritize Common Storage Areas First}: Check typical locations (like countertops or stove burners) mentioned in the task before deep cabinet search.
\end{quote}

\paragraph{WebShop.}
\begin{quote}\small
1.\ Formulate search queries with all key user specifications (e.g., ``gluten free chai orca spice flavor under \$20'').

2.\ If initial results are irrelevant, refine search terms by adding or rephrasing keywords.

3.\ Before selecting a product, cross-check all attributes (price, features, specs) against user requirements.

4.\ Use systematic navigation: proceed through pages methodically and backtrack if no matches are found.

5.\ Compare multiple products on the same page to identify the best match before clicking.
\end{quote}

\paragraph{Emergent bottleneck-category correspondence.}
Inspecting the distilled math and agentic insights above (and across the full corpora), their content maps cleanly onto the same small set of recurring bottleneck \emph{types}, even though the distillation prompts (\S\ref{app:insight_prompts}) never mention any category scheme.
Table~\ref{tab:bottleneck_categories} makes this post-hoc correspondence explicit: each row pairs a math state~$\rightarrow$~rule with an analogous agent insight, and reading it left-to-right shows the same operation, namely that at a mid-progress state, we retrieve the rule that resolves the current bottleneck.
This supports the design claim in \S\ref{sec:design_claim} that the analogy is an emergent property of the distilled insights, not an artifact engineered into the pipeline.

\begin{table*}[t]
\centering
\small
\begin{tabular}{p{2.6cm}p{5.6cm}p{5.6cm}}
\toprule
\textbf{Bottleneck category} & \textbf{Math: state $\rightarrow$ rule} & \textbf{Analogous agent insight} \\
\midrule
Search / Identify & Identify whether two triangles are similar before using ratios, or identify fixed/forbidden objects before counting. & Prioritize task-relevant locations first: check object containers/destinations named in the task (e.g., stoveburner, sidetable). \\
\midrule
State-Transform & Convert $a \mid b$ into $b \equiv 0 \pmod{a}$, or recast a configuration into coordinates. & Select a product variant / change object state before completing the goal (heat, clean, run a device). \\
\midrule
Verification & Substitute the answer back to detect extraneous roots, or check probabilities sum to $1$. & Attribute cross-checking: before any click, verify the product matches all specified attributes (price, color, size). \\
\midrule
Error-Recovery & If direct counting overcounts, switch to inclusion--exclusion, and if a ratio method fails, switch to coordinates. & Progressive filter strategy: when initial search yields partial matches, add the missing requirements to subsequent queries. \\
\midrule
Structure / Layout & Map positions onto coordinates or parity classes, or arrange constrained objects relative to one another. & Reason about the scene's spatial/containment structure: which receptacle holds the target, where the item sits in the variant grid. \\
\midrule
Strategy-before-Execute & First decide divisibility vs.\ modular vs.\ gcd/lcm, choose factoring vs.\ substitution, and plan sub-step order before executing. & Precision-first search: include all critical attributes (type, color, size, price) in the initial query, then classify the current phase and choose a plan before acting. \\
\bottomrule
\end{tabular}
\caption{Emergent correspondence between math and agentic insights across six recurring bottleneck categories. The categories are a post-hoc observation over distilled insights, not a label scheme imposed during generation, and each row instantiates the same situation $\rightarrow$ bottleneck-resolving-rule operation.}
\label{tab:bottleneck_categories}
\end{table*}

\subsection{Insight Corpus Statistics}
\label{app:insight_stats}

Main experiments (\S\ref{sec:experiments}, \S\ref{sec:results}) use DeepSeek-R1~\citep{deepseekr1} generated insight corpora for ALFWorld and WebShop, and DeepSeek-V4-Flash generated insights for ScienceWorld, each at two granularities: \textsc{Bundle} (multi-rule summaries with chain-of-thought) and \textsc{Atomic} (single rules split from bundles).
Table~\ref{tab:insight_stats} reports corpus size and length statistics for each environment, and these corpora are referenced throughout the dynamic-agent evaluation.

\begin{table*}[h]
\centering
\small
\begin{tabular}{llrrr}
\toprule
\textbf{Environment} & \textbf{Corpus} & \textbf{\# Insights} & \textbf{Avg Words} & \textbf{Range} \\
\midrule
\multirow{2}{*}{ALFWorld}
& \textsc{Bundle} & 2{,}419 & 494 & 90--1{,}114 \\
& \textsc{Atomic} & 21{,}916 & 25 & 1--365 \\
\midrule
\multirow{2}{*}{WebShop}
& \textsc{Bundle} & 501 & 549 & 176--1{,}064 \\
& \textsc{Atomic} & 5{,}212 & 24 & 1--371 \\
\midrule
\multirow{2}{*}{ScienceWorld}
& \textsc{Bundle} & 700 & 432 & 41--889 \\
& \textsc{Atomic} & 6{,}265 & 46 & 2--634 \\
\bottomrule
\end{tabular}
\caption{Insight corpus statistics (main experiments). ALFWorld and WebShop insights are generated by DeepSeek-R1~\citep{deepseekr1}, and ScienceWorld insights are generated by DeepSeek-V4-Flash. \textsc{Bundle} = multi-rule insights, \textsc{Atomic} = individual rules extracted from bundles. Range = min--max words per insight.}
\label{tab:insight_stats}
\end{table*}

\subsection{Alternative Insight-Generator Robustness}
\label{app:gpt52_topk}

\subsubsection{Corpus Statistics}
\label{app:gpt52_insight_stats}

Table~\ref{tab:gpt52_insight_stats} summarizes GPT-5.2~\citep{openai2025gpt52}-generated insight corpora used in the robustness check (\S\ref{sec:gpt52_insights}).
\textsc{Bundle} files contain one multi-rule summary per past trajectories subset, and \textsc{Atomic} files expand each bundle into individual rules.
Compared with the DeepSeek-R1~\citep{deepseekr1} corpora in Table~\ref{tab:insight_stats}, GPT-5.2 bundles are shorter on average while atomic rules are similar in length.

\begin{table*}[h]
\centering
\small
\begin{tabular}{llrrr}
\toprule
\textbf{Environment} & \textbf{Corpus} & \textbf{\# Insights} & \textbf{Avg Words} & \textbf{Range} \\
\midrule
\multirow{2}{*}{ALFWorld}
& \textsc{Bundle} & 700 & 314 & 178--500 \\
& \textsc{Atomic} & 8{,}410 & 25 & 11--58 \\
\midrule
\multirow{2}{*}{WebShop}
& \textsc{Bundle} & 700 & 358 & 198--532 \\
& \textsc{Atomic} & 10{,}162 & 23 & 4--54 \\
\bottomrule
\end{tabular}
\caption{GPT-5.2 insight corpus statistics. Range = min--max words per non-empty insight, and WebShop has 10 empty bundle summaries (word count 0). Atomic = single rules split from bundles.}
\label{tab:gpt52_insight_stats}
\end{table*}

\subsubsection{Top-$k$ Scaling Details}
\label{app:gpt52_topk_figures}

\begin{figure}[H]
\centering
\includegraphics[width=\columnwidth]{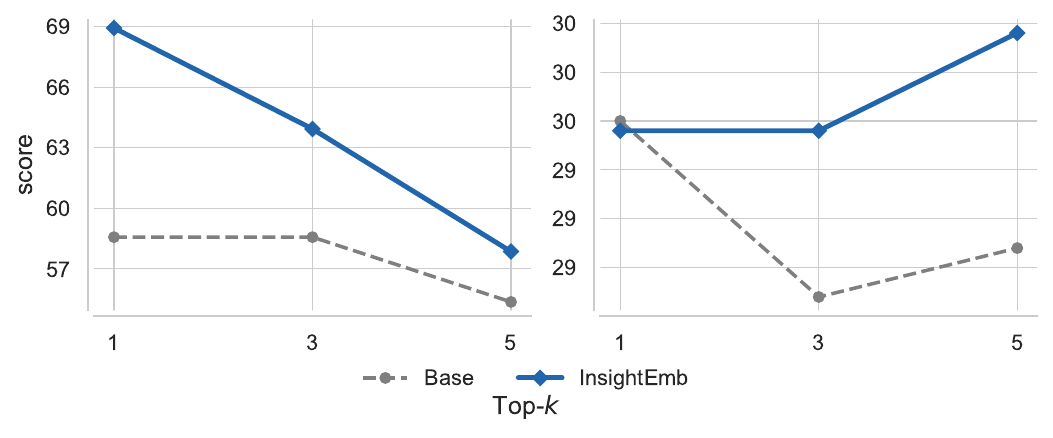}
\caption{Robustness check with GPT-5.2-generated insight corpora. \textbf{Left:} ALFWorld. \textbf{Right:} WebShop. Curves average \textsc{Bundle} and \textsc{Atomic} at each $k$ for Base and InsightEmb.}
\label{fig:gpt52_insights_corpora_average}
\end{figure}

\begin{figure}[H]
\centering
\includegraphics[width=\columnwidth]{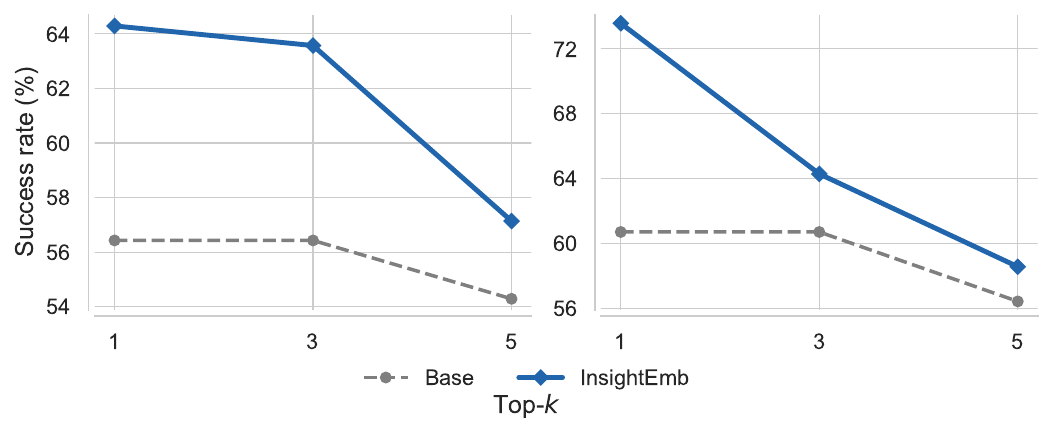}
\caption{ALFWorld success rate  vs.\ $k \in \{1,3,5\}$ with GPT-5.2-generated insights.}
\label{fig:gpt52_insights_alfworld}
\end{figure}

\begin{figure}[H]
\centering
\includegraphics[width=\columnwidth]{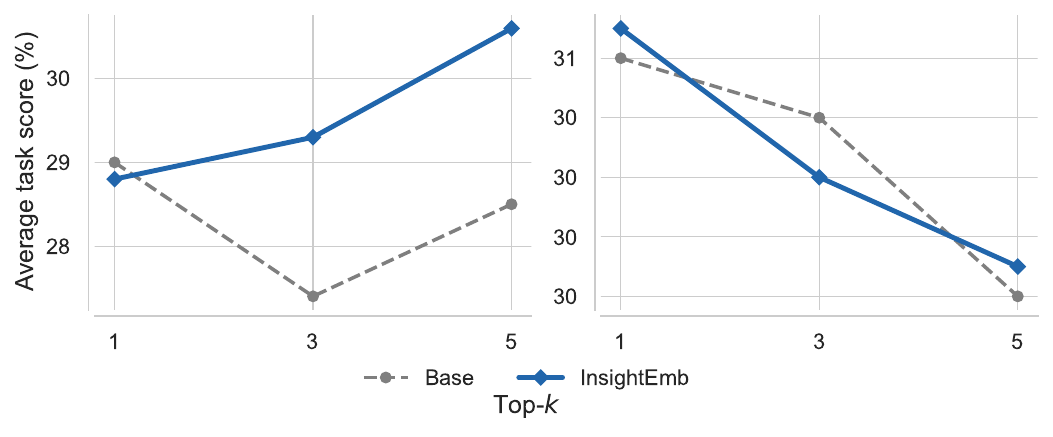}
\caption{WebShop average task score  vs.\ $k \in \{1,3,5\}$ with GPT-5.2-generated insights.}
\label{fig:gpt52_insights_webshop}
\end{figure}

Figure~\ref{fig:gpt52_insights_corpora_average} averages \textsc{Bundle} and \textsc{Atomic} at each $k$ (Qwen3-8B agent, GPT-5.2-generated insights), and InsightEmb consistently outperforms Base on both environments, confirming the improvement persists under an alternative insight generator.
Figures~\ref{fig:gpt52_insights_alfworld} and~\ref{fig:gpt52_insights_webshop} show the per-corpus breakdown.

\section{Training Setup and Hyperparameters}
\label{app:setup_hyperparams}

\subsection{Training Setup}
\label{app:training_setup}
\label{sec:training}

We fine-tune Qwen3-Embedding-4B~\citep{qwen3embedding}, building on contrastive embedding practice~\citep{wang2022text,su2023instructor}, using DeepSpeed ZeRO-3 on 8 GPUs.
Both stages use learning rate $1{\times}10^{-5}$ with cosine scheduling, per-device batch size~8 with 2 gradient accumulation steps, temperature $\tau{=}0.01$, training group size~11, maximum passage length 1{,}536 tokens, 6 epochs, and warmup ratio~0.1.
Stage~2 initializes from the Stage~1 checkpoint.
Table~\ref{tab:hyperparams} lists the full configuration.

\subsection{Hyperparameters}
\label{app:hyperparams}

Both training stages share the settings in Appendix~\ref{app:training_setup}.
Table~\ref{tab:hyperparams} lists the full configuration.

\begin{table}[h]
\centering
\small
\begin{tabular}{lcc}
\toprule
\textbf{Hyperparameter} &   \\
\midrule
Base model & Qwen3-Emb-4B \\
Learning rate & $1 \times 10^{-5}$ \\
LR scheduler & Cosine  \\
Warmup ratio & 0.1  \\
Batch size (per device) & 8  \\
Grad. accumulation & 2  \\
Training epochs & 6  \\
Temperature $\tau$ & 0.01  \\
Training group size & 11  \\
Max passage length & 1,536 \\
Precision & BF16  \\
GPUs & 8  \\
DeepSpeed & ZeRO-3 \\
Seed & 3407\\
\bottomrule
\end{tabular}
\caption{Training hyperparameters for both stages.}
\label{tab:hyperparams}
\end{table}

\section{Reasoning-Oriented Retrievers and Supervision Diversity}
\label{app:reasoning_diversity}

\subsection{Reasoning-Oriented Retriever Top-$k$ Details}
\label{app:reasoning_topk}

\begin{figure}[H]
\centering
\includegraphics[width=\columnwidth]{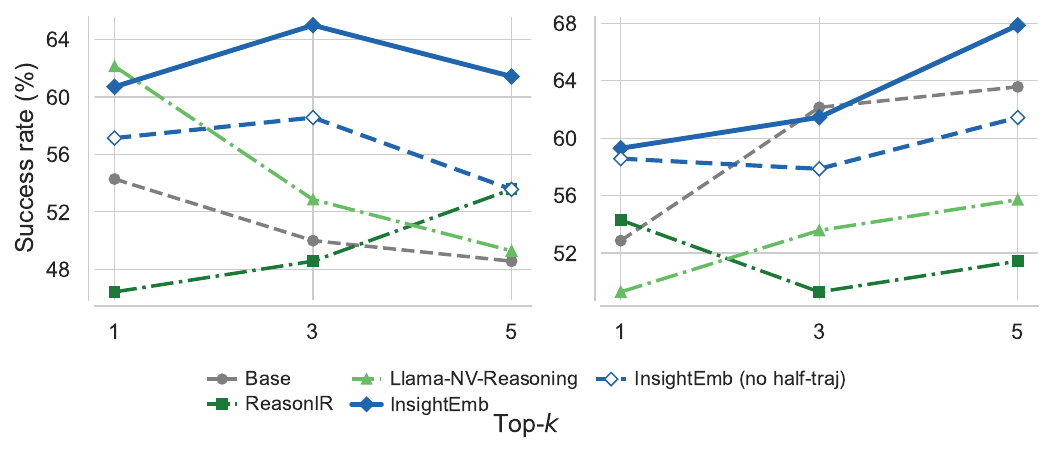}
\caption{ALFWorld test success rate  vs.\ retrieval budget $k \in \{1,3,5\}$ for Base, ReasonIR, Llama-NV-Reasoning, InsightEmb, and the \emph{no half-traj} ablation (dashed, InsightEmb color).}
\label{fig:reasoning_baselines_alfworld}
\end{figure}

\begin{figure}[H]
\centering
\includegraphics[width=\columnwidth]{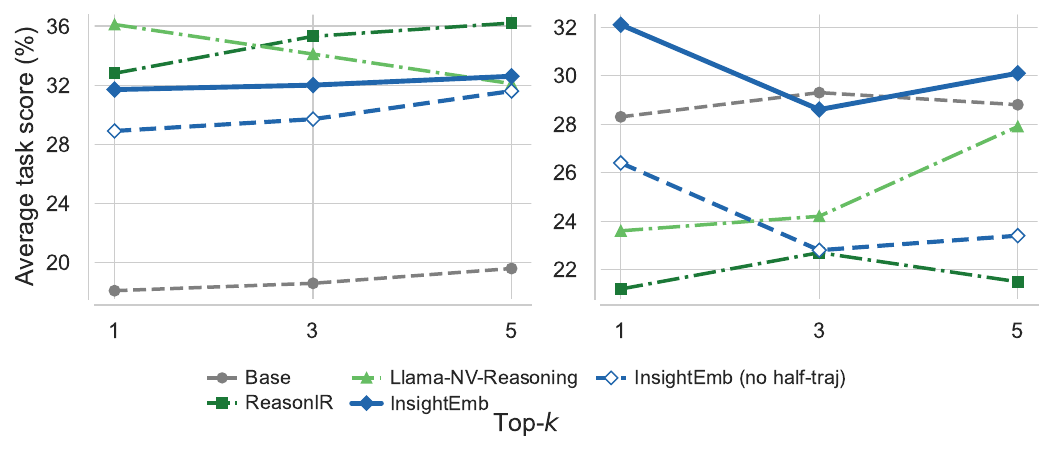}
\caption{WebShop average task score  at each $k$ for the same models and corpora.}
\label{fig:reasoning_baselines_webshop}
\end{figure}

Figure~\ref{fig:reasoning_baselines_corpora_average} in the main text averages \textsc{Bundle} and \textsc{Atomic} at each $k \in \{1,3,5\}$.
Figures~\ref{fig:reasoning_baselines_alfworld} and~\ref{fig:reasoning_baselines_webshop} show the per-corpus breakdown.
The dashed \emph{no half-traj} curve (InsightEmb color) removes partial-trajectory queries from Stage~1 while keeping the same $I^+$ and $I^-$ sets.

\subsubsection{Effect of Partial-Trajectory Queries}
\label{app:no_halftraj}

The \emph{no half-traj} ablation removes pairs whose query $q$ is a partial (incomplete) solution trace, leaving query-only and full-trajectory anchors.
The dashed curve in Figure~\ref{fig:reasoning_baselines_corpora_average} lies between Base and full InsightEmb on both environments.
At the $k$ that maximizes corpus-averaged InsightEmb on each environment, the step from \emph{no half-traj} to full InsightEmb accounts for about $83\%$ of the total Base-to-InsightEmb gain on ALFWorld and about $49\%$ on WebShop, indicating that much of InsightEmb's gains over Base come from partial-trajectory situation anchors with shared $I^+$ and $I^-$.
This also helps explain why InsightEmb outperforms static reasoning retrievers that are not trained on such mid-inference anchors.

\subsection{Why Math Training Outperforms In-Domain Fine-Tuning}
\label{app:math_vs_indomain}

As shown in \S\ref{sec:indomain} in Table~\ref{tab:indomain_alf}, the ALFWorld-trained retriever underperforms math-trained InsightEmb despite task-matched supervision.
The intuition is that ALFWorld is a comparatively narrow source: its tasks are built from a small set of recurring procedural patterns (the standard ALFWorld task types: \emph{find}, \emph{go-to}, \emph{take}, \emph{heat}/\emph{clean}/\emph{cool}, and \emph{place}), so its situation-to-rule pairs concentrate on a few structures.
Math traces, by contrast, span counting \& probability, number theory, and geometry, yielding a wider variety of situation-to-rule mappings.
We make this precise and measurable below, replacing the informal phrase ``structural diversity'' used in earlier drafts.

\subsubsection{Quantifying Structural Diversity}
\label{app:diversity_metrics}

For each situation-to-rule pair we form a normalized pair embedding $p = \mathrm{normalize}([s, r, r-s])$ with InsightEmb, where $s$ and $r$ are the normalized situation and rule embeddings and $r-s$ is their difference vector.
Within each source we randomly draw $2{,}000$ sampled pairs and compute three geometry statistics:
\begin{itemize}[leftmargin=*,nosep]
\item \textbf{Mean pairwise cosine distance}: the average of $1-\cos(p_i,p_j)$ over $5{,}000$ random pair-pair comparisons among the sampled pairs. Larger means the pairs are more spread out.
\item \textbf{Mean nearest-neighbor distance}: the average of $1-\max_{j\neq i}\cos(p_i,p_j)$ over all sampled pairs, i.e.\ each pair's cosine distance to its single closest neighbor. Larger means pairs sit farther from their nearest neighbor.
\item \textbf{Covariance effective rank}: $\exp(H)$, where $H=-\sum_k q_k\log q_k$ is the entropy of the normalized singular-value spectrum $q_k=\sigma_k^2/\sum\sigma^2$ of the covariance of the centered sampled pair embeddings. Larger means the embeddings occupy more independent directions.
\end{itemize}

On all three statistics, math pairs are more spread out than ALFWorld pairs (Table~\ref{tab:diversity}).

\begin{table}[H]
\centering
\small
\begin{tabular}{lcc}
\toprule
\textbf{Diversity metric} & \textbf{ALFWorld} & \textbf{Math} \\
\midrule
Mean pairwise cosine distance & 0.1065 & \textbf{0.2446} \\
Mean nearest-neighbor distance & 0.0372 & \textbf{0.0546} \\
Covariance effective rank & 93.27 & \textbf{100.50} \\
\bottomrule
\end{tabular}
\caption{Geometric diversity of situation-to-rule pair embeddings $[s,r,r-s]$ under InsightEmb ($2{,}000$ sampled pairs per source). Math pairs occupy a broader, higher-rank region and sit farther from their nearest neighbors than ALFWorld pairs, consistent with the narrow set of recurring ALFWorld task patterns.}
\label{tab:diversity}
\end{table}

Math situation-to-rule pairs occupy a broader, higher-rank region of the embedding geometry and are farther from their nearest neighbors, whereas ALFWorld pairs are more concentrated.
A retriever trained on the broader source therefore generalizes to bottleneck configurations that are rare or absent in ALFWorld's own games.
This is also consistent with the literature, where math reasoning is a recognized springboard for cross-domain transfer: reasoning trained on math/code with verifiable rewards yields general reasoning behaviors that transfer to STEM and other tasks~\citep{deepseekr1}, math-only training generalizes across scientific QA, agent planning, and coding~\citep{huan2025mathgeneral}, and math-first elicitation seeds broad multi-domain reasoning~\citep{pang2025bootstrapmath}.
Our contribution brings this recognized property to the \emph{retrieval} side.

\subsubsection{Transfer Despite Different Step Granularities}
\label{app:step_transfer}

The transfer does not require math and embodied tasks to share the same notion of a ``step'', relying only on the shared situation-to-rule structure (\S\ref{sec:design_claim}, Table~\ref{tab:bottleneck_categories}).
We do impose an explicit step decomposition on math traces: a chain-of-thought solution is not one atomic block, and we take \emph{prefixes} of the derivation to form partial-trajectory queries (query form (iii), \S\ref{sec:stage1}), so a single problem yields several situation-to-rule pairs at different points of progress, the math counterpart of an agent's mid-episode state.
At inference we do not require matching step boundaries: the agent's state (task + action history + observation) is encoded with a domain-neutral instruction prefix (\S\ref{sec:inference}) into the \emph{same} embedding space and matched to the nearest rule.
What crosses domains is the geometry and its shared bottleneck categories, not a step schema.
Two empirical checks confirm this despite the different step granularities. First, the \emph{no half-traj} ablation removes exactly the partial (mid-progress) math anchors and erases most of the agentic gain, showing it is the mid-progress math \emph{states} that teach the transferable property. Second, the transfer is bidirectional (Table~\ref{tab:math_insight}): an ALFWorld-trained model, whose ``steps'' are embodied actions, improves math insight retrieval on all math domains, which could not occur if the two step structures were genuinely incompatible for this purpose. The reverse direction is weaker than math-to-ALFWorld, exactly as expected if ALFWorld data covers a narrower slice of the shared geometry.

\section{Per-Environment Analysis}
\label{app:per_env}

\subsection{Per-Game-Level Agent Statistics}
\label{app:episode_stats}

This appendix supplements the aggregate results in Table~\ref{tab:main_results} with \emph{per-game-level} analysis: for each ALFWorld or WebShop test game, we log the final task score, the number of environment steps until termination, and how many \emph{distinct} insight blocks were injected over the course of that game.
All numbers use Qwen3-8B with top-1 \emph{dynamic} insight retrieval on \textsc{Bundle} summary insights, the same protocol as the \textsc{Bundle} rows in Table~\ref{tab:main_results}.
The goal is to show that InsightEmb's gains are not only a higher mean success rate, but also more varied retrieval within each game, shorter successful trajectories, and fewer complete failures.

\subsubsection{Metrics and computation}
\label{app:episode_metrics}

Each evaluated game is one independent rollout from the initial observation until success, failure, or the 50-step cap.
At every step, the current agent state is embedded and the top-1 insight is retrieved from the pre-encoded \textsc{Bundle} corpus, and that text block is prepended to the action prompt.

\textbf{Steps} counts the environment steps in that rollout.

\textbf{Distinct insights per game} counts how many \emph{different} insight strings were injected over the inference.
Retrieval is recomputed at \emph{every} step (up to 50), so each step could return a different top-1 neighbor, the count is therefore bounded by the number of steps ($\leq 50$).
The statistic measures how often the top-1 neighbor \emph{changes}, not corpus size.

\textbf{Avg.\ task score} is the mean of per-game-level outcomes: on ALFWorld and ScienceWorld, $1$ for a win and $0$ otherwise (equivalent to success rate), and on WebShop, the environment's $\texttt{task\_score}\in[0,1]$ converted to a percentage.

\textbf{Failed games} are rollouts with no positive reward. ALFWorld and WebShop task scores are bounded below by $0$, so a failed rollout is a zero-score game ($\texttt{task\_score}=0$), whereas ScienceWorld task scores can be \emph{negative} (the environment penalizes some invalid trajectories), so failed ScienceWorld rollouts are zero-or-negative-score. We report all three under a single \textbf{failed-game} count, with the \textbf{failed-game rate} equal to this count divided by the number of games.

\textbf{Retrieval turnover} is the fraction of decision steps at which the top-1 retrieved insight differs from the previous step, a step-normalized measure of how often retrieval changes as the state evolves.

\textbf{Games with $>$1 distinct insight} is the count (and rate) of games where retrieval changed at least once.

\begin{table*}[t]
\centering
\footnotesize
\setlength{\tabcolsep}{5pt}
\begin{tabular}{lcccccc}
\toprule
& \multicolumn{2}{c}{\textbf{ALFWorld}} & \multicolumn{2}{c}{\textbf{WebShop}} & \multicolumn{2}{c}{\textbf{ScienceWorld}} \\
\cmidrule(lr){2-3}\cmidrule(lr){4-5}\cmidrule(lr){6-7}
\textbf{Metric} & \textbf{Base} & \textbf{InsightEmb} & \textbf{Base} & \textbf{InsightEmb} & \textbf{Base} & \textbf{InsightEmb} \\
\midrule
\# games & 140 & 140 & 500 & 500 & 500 & 500 \\
\# failed games & 64 & 55 & 309 & 258 & 277 & 260 \\
Failed-game rate & 45.7\% & 39.3\% & 61.8\% & 51.6\% & 55.4\% & 52.0\% \\
Avg.\ task score & 54.3\% & 60.7\% & 18.42\% & 31.74\% & 7.40\% & 8.00\% \\
Avg.\ steps & 30.3 & 28.2 & 40.8 & 33.9 & 40.96 & 43.59 \\
Median steps & 30 & 24 & 50 & 50 & 7.0 & 7.0 \\
Distinct insights / game (mean) & 1.82 & 5.79 & 1.85 & 5.98 & 2.04 & 2.37 \\
Max distinct insights / game & 4 & 22 & 4 & 18 & 7 & 7 \\
\bottomrule
\end{tabular}
\caption{Per-game-level statistics (Qwen3-8B, top-1 dynamic retrieval). ALFWorld, WebShop, and ScienceWorld all use \textsc{Bundle} insights over the seed-42 balanced 500-instance subset. A \emph{failed game} is a rollout with no positive reward: for ALFWorld and WebShop, whose scores are bounded below by $0$, this is a zero-score game, whereas ScienceWorld task scores can be negative, so failed ScienceWorld games are zero-or-negative-score. \emph{Retrieval turnover} is the fraction of steps at which the top-1 neighbor changes. Failed ALFWorld/WebShop games typically reach the 50-step cap, whereas ScienceWorld rollouts terminate earlier on average (median 7 steps).}
\label{tab:episode_stats}
\end{table*}

\subsubsection{ALFWorld (140 test games)}
\label{app:episode_alf}

On ALFWorld, InsightEmb improves every column in Table~\ref{tab:episode_stats}.

\paragraph{Retrieval turnover.} Retrieval runs at every step, so distinct insights per game is bounded by steps ($\leq 50$).
Base nevertheless uses only 1.82 distinct strings on average (max~4) while InsightEmb averages 5.79 (max~22).
The comparison is about how often top-1 changes as the state evolves, not about injecting a new corpus block every step.
\paragraph{Efficiency.} Mean steps decrease from 30.3 to 28.2 and the \emph{median} drops from 30 to 24, so at least half of InsightEmb games finish in $\leq$24 steps whereas Base's median sits at the cap for unsuccessful search-heavy runs.
Shorter medians co-occur with higher success: InsightEmb avoids prolonged cabinet-by-cabinet loops documented in Appendix~\ref{app:qualitative_example}.

\paragraph{Failures.} Zero-score games (failed rollouts) decrease from 64 to 55, showing gains even in completely failed games.

\subsubsection{WebShop (500 test games)}
\label{app:episode_webshop}

\paragraph{Retrieval turnover.} With per-step top-1 retrieval over up to 50 steps, distinct insights per game could be as large as the step count, and Base averages 1.85 (max~4) while InsightEmb averages 5.98 (max~18).
InsightEmb changes the retrieved block in 487/500 games ($97\%$) vs.\ 295 ($59\%$) for Base, so different steps can surface search, variant-selection, and checkout rules from the \textsc{Bundle} corpus.

\paragraph{Efficiency.} Average steps drop from 40.8 to 33.9 (mean rollout ${\sim}$41 steps) while both medians remain at 50, because many games still hit the cap, so InsightEmb shortens trajectories it completes rather than shifting the median below the horizon.
Appendix~\ref{app:webshop_examples} illustrates loop-heavy Base runs vs.\ earlier termination when purchase rules are retrieved.

\paragraph{Failures.} Zero-score games decrease from 309 to 258 ($-51$), the gain comes from converting full failures into partial or full purchase credit.

\subsubsection{ScienceWorld (500 test games)}
\label{app:episode_scienceworld}

ScienceWorld statistics in Table~\ref{tab:episode_stats} use the \textsc{Bundle} corpus and the seed-42 balanced 500-instance subset (top-1, Qwen3-8B).

\paragraph{Failures.} Negative-score games decrease from 277 to 260 ($-17$), lowering the negative-score rate from $55.4\%$ to $52.0\%$, so InsightEmb reduces complete failures even on this hard environment, consistent with its small but positive success gain ($7.40\% \rightarrow 8.00\%$).

\paragraph{Retrieval turnover.} On ScienceWorld the retrieval-diversity pattern differs from ALFWorld and WebShop. When normalized by trajectory length, retrieval turnover \emph{rises} from $0.268$ to $0.380$ and the mean distinct insights per game from $2.04$ to $2.37$ (both medians at $2.0$, max $7$ for both), so InsightEmb still updates its top-1 neighbor more often \emph{per step} as the experimental state evolves.
At the same time, the raw count of games with more than one distinct insight \emph{falls} from $330/500$ ($66.0\%$) for Base to $224/500$ ($44.8\%$) for InsightEmb.
These are consistent rather than contradictory: InsightEmb more often locks onto a single procedurally complete workflow insight and reuses it across many steps of an experiment (\S\ref{app:scienceworld_step_dynamics}), so within a game it changes insights less often overall, but the changes it does make track the current sub-procedure more tightly. In other words, ScienceWorld rewards \emph{procedural completeness} of a retrieved insight over per-observation novelty, unlike the search-heavy ALFWorld/WebShop trajectories where higher raw turnover co-occurs with success.

\paragraph{Efficiency.} The median rollout length is short ($7$ steps for both), reflecting many quickly-terminating games, and InsightEmb's higher \emph{mean} steps ($40.96 \rightarrow 43.59$) indicate it more often persists through the multi-step experimental protocol rather than terminating early without completing the required state transition.

\subsection{ALFWorld Per-Task and Mechanistic Analysis}
\label{app:alfworld_analysis}

This section expands the ALFWorld mechanistic claims summarized in \S\ref{sec:results}: per-task-type breakdown, multi-granularity invariance, qualitative retrieval examples, rule-level topical vs.\ procedural matching, and step-conditioned retrieval dynamics.

\subsubsection{Per-Task-Type Breakdown}
\label{app:per_task}

Table~\ref{tab:per_task} supplements the aggregate ALFWorld results in Table~\ref{tab:main_results} by reporting test success counts for each ALFWorld task family (\textsc{Atomic} corpus, top-1 retrieval, Qwen3-8B).
Counts are out of the games listed in parentheses per type, and bold marks the higher count between Base and InsightEmb.

\begin{table}[h]
\centering
\small
\begin{tabular}{lcc}
\toprule
\textbf{Task Type} & \textbf{Base} & \textbf{Ours} \\
\midrule
clean (22) & 27 & \textbf{41} \\
cool (26) & 62 & \textbf{69} \\
heat (19) & 42 & \textbf{53} \\
put (48) & 71 & 71 \\
examine (17) & \textbf{71} & 53 \\
find\_two (8) & 12 & \textbf{38} \\
\midrule
\textbf{Overall} & 55.0 & \textbf{59.3} \\
\bottomrule
\end{tabular}
\caption{ALFWorld success rates by task type (top-1 retrieval, Qwen3-8B). Base = \textsc{Atomic}/Base, and Ours = \textsc{Atomic}/InsightEmb. Numbers in parentheses indicate total games per type.}
\label{tab:per_task}
\end{table}

\subsubsection{Multi-Granularity Query Invariance}
\label{app:granularity}

Our Stage~1 design enforces that query-only, partial-trajectory, and full-trajectory versions of the same problem all retrieve the same insight.
This directly maps to the agentic setting: at step~0 the agent has only the task description (analogous to query-only), at mid-game it has partial action history (partial trajectory), and at late steps it has extensive history (full trajectory).
The multi-granularity training ensures that the correct insight is retrievable at every stage of task execution, not just at the beginning.

\subsubsection{Qualitative Example: Insight-Guided Search}
\label{app:qualitative_example}

Figure~\ref{fig:qualitative} illustrates how InsightEmb's retrieved insights lead to better action selection on a representative task (``put a hot potato in fridge'').

\begin{figure}[H]
\centering
\fbox{\parbox{0.95\columnwidth}{\small
\textbf{Task}: ``put a hot potato in fridge''
\\[4pt]
\textbf{Base model retrieves} (\textsc{Atomic}/Base):
\\\emph{``Apply Task-Specific Physics Reasoning: Heat transfers require closed appliances. Objects in containers remain inaccessible until explicit retrieval.''}
\\$\rightarrow$ Agent focuses on heating procedure but cannot locate the potato
\\$\rightarrow$ \textbf{Timeout at 50 steps}
\\[4pt]
\textbf{InsightEmb retrieves} (\textsc{Atomic}/InsightEmb):
\\\emph{``Prioritize Task-Relevant Locations First: Always check object containers/destinations mentioned in the task (e.g., stoveburner, sidetable).''}
\\$\rightarrow$ Agent locates potato $\rightarrow$ heat $\rightarrow$ place in fridge
\\$\rightarrow$ \textbf{Success in 37 steps}
}}
\caption{Qualitative comparison on Game~0 (test, top-5 retrieval). The base model's insight describes the heating \emph{procedure} but not \emph{where to find} the potato, whereas InsightEmb retrieves a search-priority rule that addresses the actual bottleneck.}
\label{fig:qualitative}
\end{figure}

The base model's insight is topically relevant but strategically vacuous, as it does not indicate \emph{where} to find the potato.
InsightEmb retrieves a structurally matched search-priority rule, directly determining the agent's first action.
This pattern of generic vs.\ structurally specific retrieval recurs across the 16 games that InsightEmb wins uniquely under top-1 retrieval (see also \S\ref{app:qualitative}).

\subsubsection{Topical vs.\ Procedural Retrieval: Rule-Level Analysis}
\label{app:rule_analysis}

Because each \textsc{Atomic} insight is a single rule, we can examine which \emph{type} of rule each embedding model prioritizes for the same query.
We classify retrieved rules with GPT-5.2~\citep{openai2025gpt52} into functional categories: \textsc{Where} (search strategy), \textsc{How} (state-change procedures), \textsc{Place} (destination logic), and \textsc{Verify} (state checks).

The Base model retrieves rules about \emph{how to heat} (``Apply Task-Specific Physics Reasoning,'' ``Apply Thermodynamic Context Filtering,'' ``Appliance Function Mapping''), which describe the heating \emph{procedure} but not \emph{where to find the potato}.
InsightEmb retrieves rules about \emph{where to search} (``Prioritize Task-Relevant Locations First,'' ``Prioritize Direct Affordances,'' ``Prioritize Immediate Target Interaction''), which address the actual bottleneck: the agent's first action must be to \emph{locate} the target object.

This reveals a key difference in how the two models understand task structure:
\begin{itemize}[leftmargin=*,nosep]
    \item \textbf{Base} matches the query to rules that share the task's \emph{topic} (heating $\rightarrow$ heating rules). This is \emph{topical matching}.
    \item \textbf{InsightEmb} matches the query to rules that address the task's \emph{current bottleneck} (the agent hasn't found the object yet $\rightarrow$ search rules). This is \emph{action-intent matching}: retrieving the rule implied by what the agent needs to do next.
\end{itemize}

The action-intent match is correct: in ALFWorld, the agent must first \emph{find} the target object before it can apply any state transformation.
InsightEmb has learned this sequential dependency from the structural parallel in math, where a problem must first be \emph{understood} (matched to the right strategy) before it can be \emph{solved} (executed step by step).

\subsection{Step-Conditioned Retrieval Details}
\label{app:step_dynamics}

\begin{figure}[h]
\centering
\includegraphics[width=0.48\textwidth]{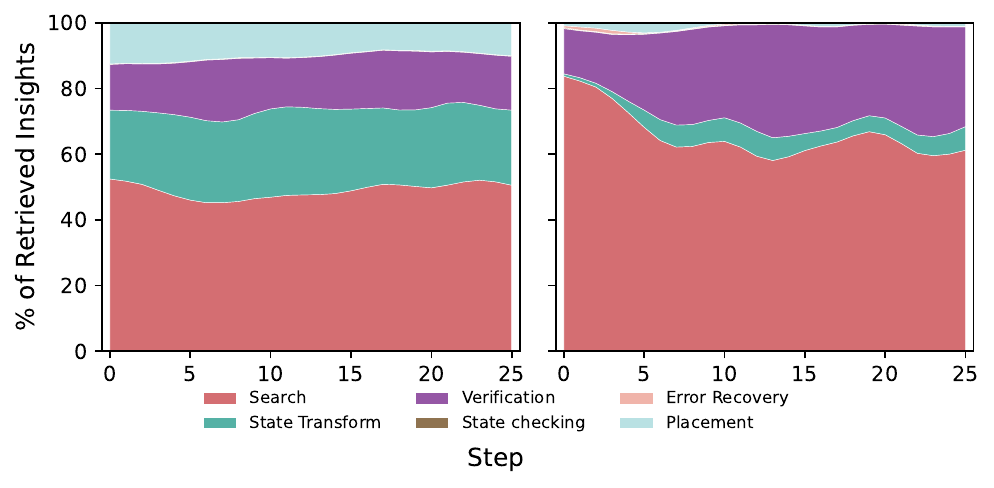}
\caption{Per-step retrieved insight category distribution (\textsc{Atomic}, test, top-1 dynamic retrieval, 3-step rolling average). \textbf{Left:} Base embedder. \textbf{Right:} InsightEmb. Base changes little across steps, whereas InsightEmb shifts from Search toward Verification and State Transform.}
\label{fig:step_area}
\end{figure}

The rule-level analysis above examines \emph{which} rules are retrieved, and we now examine \emph{when} they are retrieved.
We extract the top-1 retrieved \textsc{Atomic} insight at every step of all 140 test games, classify each rule with GPT-5.2~\citep{openai2025gpt52}, and plot the per-step category mix in Figure~\ref{fig:step_area} with 3-step rolling average.
For readability we merge \textsc{Search \& Locate} with \textsc{Navigation} into \textbf{Search}, and \textsc{State Tracking} with \textsc{Task Planning} into \textbf{State checking} (six categories total, with definitions and examples below).

\paragraph{Base vs.\ InsightEmb dynamics.}
Base shows almost no step-wise change: \textbf{Search} is 49.7\% on average over steps 1--4 vs.\ 51.0\% over steps 20--25 ($\Delta{=}{+}1.3$ pp), \textbf{State Transform} 23.0\% $\rightarrow$ 23.6\%, and \textbf{Verification} 14.8\% $\rightarrow$ 16.1\%.
InsightEmb instead shows a clear \emph{state-aware} shift along the episode: \textbf{Search} falls from 78.1\% (steps 1--4) to 61.7\% (steps 20--25, $\Delta{=}{-}16.4$ pp), while \textbf{Verification} rises from 16.9\% to 31.5\% ($\Delta{=}{+}14.6$ pp) and \textbf{State Transform} from 1.9\% to 5.9\% ($\Delta{=}{+}3.9$ pp).
That is, InsightEmb retrieval tracks procedural phase (heavy search while the object is still missing, then more checks and state-change rules as the agent approaches completion), whereas Base remains roughly uniform.

\paragraph{Category definitions and illustrative rules.}
Each label denotes the \emph{procedural role} of a single atomic rule (not the ALFWorld task type):
\begin{description}[leftmargin=1.5em,style=nextline,itemsep=2pt]
    \item[\textbf{Search}] Rules for finding the target or reaching the right place before manipulation.
    \emph{Example:} ``Prioritize Task-Relevant Locations First: Always check object containers/destinations mentioned in the task (e.g., stoveburner, sidetable).''
    \item[\textbf{State Transform}] Rules for changing object state (clean, heat, cool) at the correct appliance.
    \emph{Example:} ``Apply Contextual Cleaning Protocol: When `clean' is specified in the task, automatically prioritize sinkbasin interaction after obtaining the target object, before any placement attempts.''
    \item[\textbf{Verification}] Rules for confirming prerequisites or completion before the next action.
    \emph{Example:} ``Ensure Cooling Completion Before Placement: Verify the cooling action (e.g., fridge interaction) is fully executed and acknowledged by the environment before attempting to move the object to its final destination.''
    \item[\textbf{State checking}] Rules for tracking progress, ordering sub-goals, or maintaining awareness of what was already searched or carried.
    \emph{Example:} ``Maintain a Dynamic Search Tracker: Systematically record searched locations and their contents to avoid redundant checks and focus exploration on unexamined areas.''
    \item[\textbf{Error Recovery}] Rules for escaping loops or reformulating strategy after failed attempts.
    \emph{Example:} ``Implement Loop Detection: Establish a threshold (3--5 attempts) for repeated unsuccessful navigation patterns, then trigger systematic re-evaluation of object location hypotheses.''
    \item[\textbf{Placement}] Rules for putting the object at its final receptacle.
    \emph{Example:} ``Container Proximity Hierarchy: When storing objects, first navigate to the nearest valid container type specified in the task (drawer/cabinet) before considering distant alternatives.''
\end{description}

\subsection{ALFWorld Qualitative Case Studies}
\label{app:qualitative}

We present two additional examples where InsightEmb succeeds while Base fails (top-1 retrieval).

\paragraph{Game 9: ``clean some ladle and put it in countertop.''}
InsightEmb checks \texttt{cabinet 1--3}, then \texttt{sinkbasin 1}, opens \texttt{cabinet 4}, finds the ladle on \texttt{countertop 1}, cleans it at the sink, and places it on the countertop, completing in \textbf{12 steps}.
Base loops through cabinets and the sink without taking the ladle, timing out at 50 steps.
This illustrates task-focused search: InsightEmb reaches the sink and visible surfaces early, while Base's generic ``check fridges first'' heuristic delays locating the ladle.

\paragraph{Game 33: ``put a cool lettuce in countertop.''}
InsightEmb opens the fridge, scans \texttt{countertop 1--3}, takes the lettuce from \texttt{countertop 3}, cools it in the fridge, and places it on the countertop, completing in \textbf{10 steps}.
Base opens the fridge but then wanders through drawers and cabinets, never retrieving the lettuce, timing out at 50 steps.
Base's retrieved summary rule prioritizes ``common storage areas'' broadly, whereas InsightEmb's rule set emphasizes checking task-mentioned surfaces (countertops) before deep cabinet search.

\subsection{WebShop Qualitative Examples}
\label{app:webshop_examples}

We present illustrative examples from the WebShop evaluation where InsightEmb with \textsc{Atomic} insights succeeds while Base fails, or achieves substantially higher task scores.
Manual inspection of divergent games (summarized in \S\ref{sec:webshop_results}) reveals three recurring qualitative patterns, which the examples below demonstrate with side-by-side Base vs.\ InsightEmb trajectories:
\begin{enumerate}[leftmargin=*,nosep]
    \item \textbf{Variant selection awareness.} InsightEmb's insights guide the agent to explicitly select product variants (color, size) before purchasing, while Base frequently skips this step, resulting in partial scores instead of perfect scores.
    \item \textbf{Loop prevention.} Base gets stuck in search--browse--back loops for 21--50 steps, whereas InsightEmb's insights about session management and error recovery help the agent break out of unproductive cycles.
    \item \textbf{Procedural sequencing.} InsightEmb retrieves insights that encode a sequential workflow (search $\rightarrow$ verify $\rightarrow$ select variants $\rightarrow$ buy), while Base retrieves topically relevant but procedurally vague insights.
\end{enumerate}
Together these show InsightEmb performs \emph{procedural matching}, retrieving insights that address the agent's current bottleneck.

\subsubsection{Example 1: Office Chair (Game~226, Atomic)}

\textbf{Task}: \emph{``Find me height adjustable, high density, easy install, easy assemble home office chairs for living room with color: type 7-pink, and price lower than 120.00 dollars.''}

\textbf{InsightEmb}: WON in 6 steps (score = 1.0). \textbf{Base}: LOST in 3 steps (score = 0.857).

\paragraph{InsightEmb insight:}
\begin{quote}\small
\emph{Progressive Filter Strategy}: When initial search yields partial matches, systematically add missing requirements to subsequent queries.
\end{quote}

\paragraph{Base insight:}
\begin{quote}\small
\emph{Precision-First Search}: Always include all critical attributes in the initial search query using AND logic for better filtering.
\end{quote}

\paragraph{Action comparison.}
Both models find the same product (\texttt{b08p8lrfz4}), but:
\begin{itemize}[leftmargin=*,nosep]
    \item \textbf{InsightEmb}: search $\rightarrow$ click product $\rightarrow$ check features $\rightarrow$ go back $\rightarrow$ \textbf{select ``type 7-pink''} $\rightarrow$ buy now \cmark
    \item \textbf{Base}: search (with AND syntax) $\rightarrow$ click product $\rightarrow$ \textbf{buy now immediately} (without selecting color variant) \xmark
\end{itemize}

\paragraph{Analysis.}
InsightEmb's ``Progressive Filter Strategy'' guides the agent to inspect product features and select the correct variant before purchasing.
Base's ``Precision-First Search'' focuses on query formulation but does not instruct variant selection, causing premature purchase with a partial score.

\subsubsection{Example 2: Women's Sweater (Game~479, Atomic)}

\textbf{Task}: \emph{``Find me women's sweaters with relaxed fit, long sleeve with color: light heather grey, and size: large, and price lower than 50.00 dollars.''}

\textbf{InsightEmb}: WON in 8 steps (score = 1.0). \textbf{Base}: LOST in 3 steps (score = 0.600).

\paragraph{InsightEmb insight:}
\begin{quote}\small
\emph{Attribute Cross-Checking}: Before any click, verify that the product matches all specified attributes (price, color, size, etc.).
\end{quote}

\paragraph{Action comparison.}
Both find the same product (\texttt{b07dkgjr74}), but:
\begin{itemize}[leftmargin=*,nosep]
    \item \textbf{InsightEmb}: search $\rightarrow$ explore first result $\rightarrow$ back to search $\rightarrow$ re-search $\rightarrow$ click product $\rightarrow$ \textbf{select ``light heather grey''} $\rightarrow$ \textbf{select ``large''} $\rightarrow$ buy now \cmark
    \item \textbf{Base}: search $\rightarrow$ click product $\rightarrow$ \textbf{buy now immediately} (without selecting color or size) \xmark
\end{itemize}

\paragraph{Analysis.}
This is the clearest example of the \emph{variant selection awareness} pattern.
InsightEmb's ``Attribute Cross-Checking'' rule explicitly guides the agent to verify and select each product variant before purchasing.
Base's generic search-formulation rule does not encode this procedural constraint, and the agent buys immediately without selecting the required color and size variants.

\subsubsection{Example 3: Dining Set (Game~287, Atomic)}

\textbf{Task}: \emph{``Find me button tufted, mid century, high density, easy assemble dining sets with solid wood, wood frame for dining room with color: light grey, and price lower than 250.00 dollars.''}

\textbf{InsightEmb}: WON in 4 steps (score = 1.0). \textbf{Base}: LOST in 50 steps (score = 0.000).

This is the most dramatic example ($\Delta = +1.000$).

\paragraph{InsightEmb insight:}
\begin{quote}\small
\emph{Precision-First Search}: Always include all critical attributes (type, color, size, price limit) in the initial search query using AND logic for better filtering.
\end{quote}

\paragraph{Action comparison.}
\begin{itemize}[leftmargin=*,nosep]
    \item \textbf{InsightEmb}: search (comprehensive query) $\rightarrow$ click product (\texttt{b09gy58gdh}) $\rightarrow$ \textbf{select ``light grey''} $\rightarrow$ buy now \cmark\ (4 steps)
    \item \textbf{Base}: search $\rightarrow$ back to search $\rightarrow$ re-search $\rightarrow$ back to search $\rightarrow$ re-search $\rightarrow$ click next $\rightarrow$ click next $\rightarrow$ back to search $\rightarrow$ \ldots\ (50 steps, never clicks a product to purchase)
\end{itemize}

\paragraph{Analysis.}
InsightEmb's concise insight produces an effective search query that finds the product immediately.
Base gets stuck in a search--refine--next--back loop for all 50 steps, illustrating the \emph{loop prevention} failure: without a clear recovery strategy, the agent exhaustively re-searches without ever committing to a product.

\subsubsection{Example 4: iPad Case (Game~190, Bundle)}

\textbf{Task}: \emph{``Find me compatible apple online game services with case cover with color: coast coconut trees, and price lower than 50.00 dollars.''}

\textbf{InsightEmb}: WON in 6 steps (score = 1.0). \textbf{Base}: LOST in 50 steps (score = 0.000).

\paragraph{Action comparison.}
\begin{itemize}[leftmargin=*,nosep]
    \item \textbf{InsightEmb}: search ``apple game case cover'' $\rightarrow$ back to search $\rightarrow$ \textbf{re-search ``iPad case coconut tree design''} $\rightarrow$ click product (\texttt{b07vvvdrcx}) $\rightarrow$ \textbf{select ``coast coconut trees''} $\rightarrow$ buy now \cmark\ (6 steps)
    \item \textbf{Base}: search ``Apple case cover'' $\rightarrow$ click next $\rightarrow$ next $\rightarrow$ next $\rightarrow$ next $\rightarrow$ back to search $\rightarrow$ re-search $\rightarrow$ next $\rightarrow$ next $\rightarrow$ next $\rightarrow$ \ldots\ (50 steps, navigating pages without clicking any product)
\end{itemize}

\paragraph{Analysis.}
InsightEmb's agent reformulates the search query after the first attempt fails, switching from the literal task wording to a more product-oriented query (``iPad case coconut tree design'').
Base's agent repeatedly paginates through results without reformulating, illustrating the \emph{loop prevention} pattern: InsightEmb's retrieval surfaces strategies for query refinement and error recovery.

\subsubsection{Example 5: Men's Sleepwear (Game~166, Atomic)}

\textbf{Task}: \emph{``Find me men's sleep \& lounge with long sleeve, elastic waistband for daily wear with color: multi 10, and size: xx-large, and price lower than 80.00 dollars.''}

\textbf{InsightEmb}: WON in 9 steps (score = 1.0). \textbf{Base}: LOST in 7 steps (score = 0.667).

\paragraph{InsightEmb insight:}
\begin{quote}\small
\emph{Compound Feature Recognition}: For combined requirements, search for the most distinctive pairing first, then check secondary attributes in product details.
\end{quote}

\paragraph{Action comparison.}
\begin{itemize}[leftmargin=*,nosep]
    \item \textbf{InsightEmb}: search $\rightarrow$ click product (\texttt{b09nd9dp7j}) $\rightarrow$ check description $\rightarrow$ back to search $\rightarrow$ re-search with exact attributes $\rightarrow$ click same product $\rightarrow$ \textbf{select ``multi 10''} $\rightarrow$ \textbf{select ``xx-large''} $\rightarrow$ buy now \cmark
    \item \textbf{Base}: search $\rightarrow$ click product (\texttt{b09nd8p2qr}) $\rightarrow$ check features $\rightarrow$ back to search $\rightarrow$ re-search $\rightarrow$ click same product $\rightarrow$ \textbf{buy now without selecting variants} \xmark
\end{itemize}

\paragraph{Analysis.}
InsightEmb's ``Compound Feature Recognition'' guides the agent through a verify-then-select workflow: first confirm the product matches, then explicitly select each variant before purchase.
Base's agent finds a similar product but purchases without selecting the required color and size, resulting in a partial score. This combines \emph{variant selection awareness} with \emph{procedural sequencing}: InsightEmb encodes a sequential workflow (search $\rightarrow$ verify $\rightarrow$ select variants $\rightarrow$ buy).

\subsubsection{Summary of WebShop Qualitative Patterns}

Across all divergent games, three consistent patterns emerge.
The qualitative cases above and the main WebShop column in Table~\ref{tab:main_results}.

\begin{enumerate}[leftmargin=*,nosep]
    \item \textbf{Variant selection awareness} (Games~226, 479, 393, 166): InsightEmb consistently guides the agent to select product variants (color, size) before purchasing. Base frequently skips this step, resulting in partial scores (0.600--0.857) instead of perfect scores. This is the single most impactful behavioral difference.
    \item \textbf{Loop prevention} (Games~287, 190): Base gets stuck in search--browse--back loops for 50 steps, scoring zero. InsightEmb's insights about query reformulation and error recovery help the agent find the product and complete the purchase.
    \item \textbf{Procedural sequencing} (Game~166): InsightEmb retrieves insights that encode a sequential workflow (search $\rightarrow$ verify $\rightarrow$ select variants $\rightarrow$ buy), while Base retrieves topically relevant but procedurally vague insights. This mirrors the ALFWorld finding where InsightEmb performs \emph{procedural matching} rather than \emph{topical matching}.
\end{enumerate}

\subsection{ScienceWorld Evaluation Details}
\label{app:scienceworld}

This appendix supplements the ScienceWorld results in Table~\ref{tab:main_results} (\S\ref{sec:scienceworld_results}) with the full top-$k$ breakdown, a stronger closed-source action model across all three environments, and the task distribution of the evaluation subset.

\subsubsection{Setup}
\label{app:scienceworld_setup}

We run ScienceWorld~\citep{wang2022scienceworld} with the identical insight-retrieval pipeline used for ALFWorld and WebShop: Qwen3-8B as the action-generating LLM with greedy decoding, a history window of 3 steps, and dynamic top-$k$ insight retrieval over the ScienceWorld insight corpus (700 \textsc{Bundle} / 6{,}265 \textsc{Atomic} insights, Appendix Table~\ref{tab:insight_stats}).
The 500 evaluation instances are a fixed, seed-42, task-balanced subset of all ScienceWorld test variations, saved once and reused across all experiments for comparability.
We report success rate (\%). The no-insight baseline (no retrieval at all) scores only $2.40$.

\subsubsection{Top-$k$ Retrieval Budgets}
\label{app:scienceworld_topk}

Table~\ref{tab:scienceworld_topk} reports ScienceWorld success rate for Base and InsightEmb across corpus granularities and retrieval budgets $k\in\{1,3,5\}$.
InsightEmb wins or ties Base on all six cells, and every insight setting more than triples the no-insight baseline ($2.40$).
Because ScienceWorld does not add a qualitatively new insight-pool regime beyond the good/bad pools already illustrated by ALFWorld and WebShop, the main-text top-$k$ figure (Figure~\ref{fig:reasoning_baselines_corpora_average}) retains only ALFWorld and WebShop, and the ScienceWorld top-$k$ results are presented here in tabular form.

\begin{table}[H]
\centering
\small
\begin{tabular}{llcc}
\toprule
\textbf{Budget} & \textbf{Corpus} & \textbf{Base} & \textbf{InsightEmb} \\
\midrule
top-1 & \textsc{Bundle} & 7.40 & \textbf{8.00} \\
top-1 & \textsc{Atomic} & 7.40 & \textbf{10.20} \\
\midrule
top-3 & \textsc{Bundle} & 8.20 & \textbf{9.60} \\
top-3 & \textsc{Atomic} & 7.20 & \textbf{8.40} \\
\midrule
top-5 & \textsc{Bundle} & 8.80 & \textbf{10.80} \\
top-5 & \textsc{Atomic} & \textbf{9.60} & \textbf{9.60} \\
\bottomrule
\end{tabular}
\caption{ScienceWorld success rate (\%) across retrieval budgets $k\in\{1,3,5\}$ (Qwen3-8B agent, 500 seed-42 task-balanced test instances). The no-insight baseline is $2.40$, so every insight setting more than triples it. Bold marks the better value within each row, and ties are bolded for both.}
\label{tab:scienceworld_topk}
\end{table}

\subsubsection{Stronger Closed-Source Action Model (GPT-5.2)}
\label{app:scienceworld_gpt52}

To test whether the advantage depends on the Qwen3-8B action model, we replace it with a stronger, closed-source action model (GPT-5.2~\citep{openai2025gpt52}) and re-run all three environments with top-1 retrieval.
As shown in Table~\ref{tab:gpt52_cross_env}, InsightEmb improves over Base across ScienceWorld, ALFWorld, and WebShop, indicating that the gains are not tied to a specific action model.

\begin{table}[H]
\centering
\small
\begin{tabular}{llcc}
\toprule
\textbf{Environment} & \textbf{Corpus} & \textbf{Base} & \textbf{InsightEmb} \\
\midrule
\multirow{2}{*}{ScienceWorld}
& \textsc{Bundle} & 36.4 & \textbf{45.8} \\
& \textsc{Atomic} & 37.2 & \textbf{40.0} \\
\midrule
\multirow{2}{*}{ALFWorld}
& \textsc{Bundle} & 63.6 & \textbf{70.0} \\
& \textsc{Atomic} & 60.0 & \textbf{73.6} \\
\midrule
\multirow{2}{*}{WebShop}
& \textsc{Bundle} & 18.9 & \textbf{26.1} \\
& \textsc{Atomic} & 32.2 & \textbf{32.5} \\
\bottomrule
\end{tabular}
\caption{Dynamic agent evaluation with a stronger closed-source action model (GPT-5.2, top-1 retrieval) across all three environments. Reported as success rate (\%) for ScienceWorld and ALFWorld, and average task score (\%) for WebShop. Replacing the Qwen3-8B action model with GPT-5.2 preserves the InsightEmb advantage.}
\label{tab:gpt52_cross_env}
\end{table}

\subsubsection{Task Distribution of the 500-Instance Subset}
\label{app:scienceworld_task_dist}

Table~\ref{tab:scienceworld_task_dist} lists the most frequent ScienceWorld task types in the fixed seed-42 balanced subset used for all ScienceWorld experiments.

\begin{table}[H]
\centering
\small
\begin{tabular}{lc}
\toprule
\textbf{Task type} & \textbf{Count} \\
\midrule
inclined-plane-friction-named-surfaces & 37 \\
use-thermometer & 27 \\
find-animal & 26 \\
test-conductivity & 26 \\
test-conductivity-of-unknown-substances & 26 \\
find-plant & 24 \\
measure-melting-point-known-substance & 24 \\
find-non-living-thing & 23 \\
measure-melting-point-unknown-substance & 22 \\
mendelian-genetics-unknown-plant & 22 \\
\bottomrule
\end{tabular}
\caption{Top task-type counts in the fixed, seed-42, task-balanced 500-instance ScienceWorld test subset (saved once and reused across all experiments for comparability).}
\label{tab:scienceworld_task_dist}
\end{table}

\subsubsection{Task-Type Grouping}
\label{app:scienceworld_per_task_sec}

To parallel the ALFWorld per-task-type analysis (Table~\ref{tab:per_task}), the most frequent ScienceWorld task types (Table~\ref{tab:scienceworld_task_dist}) fall into the functional categories the agent must resolve: \emph{search/identify} (find-animal, find-plant, find-non-living-thing), \emph{measurement/verification} (use-thermometer, measure-melting-point), \emph{state-transform/experiment} (test-conductivity), and \emph{multi-step reasoning} (inclined-plane-friction, mendelian-genetics), mirroring the \textsc{Where}/\textsc{How}/\textsc{Verify} rule taxonomy used for ALFWorld (\S\ref{app:rule_analysis}).

\subsubsection{Step-Conditioned Retrieval Dynamics}
\label{app:scienceworld_step_dynamics}

Although retrieval is performed at each decision step, the most useful retrieved insights are not always narrowly tied to a single observation.
In ScienceWorld, many tasks require executing a stable multi-step procedure across changing observations.
For example, an insight such as ``to solve inclined-plane friction tasks, place the object on the ramp, vary the relevant surface or angle, observe whether it slides, then compare outcomes'' can guide several consecutive actions: setting up the apparatus, manipulating the relevant variable, observing the outcome, and making a comparison.
Thus a single high-level procedural insight may remain useful across many environment steps, even as the observation changes.

This helps explain why InsightEmb does not necessarily require more diverse or more state-specific retrievals.
Instead of retrieving a different insight for every local observation, it often retrieves a general but actionable workflow that covers the full experimental structure. In this sense the retrieved insight acts less like a one-step hint and more like a compact policy sketch.
This is consistent with Table~\ref{tab:episode_stats}, where the mean distinct retrieved insights are broadly similar between Base and InsightEmb ($2.04$ vs.\ $2.37$), while InsightEmb appears to retrieve insights that are more procedurally complete and reusable across steps (higher retrieval turnover, $0.268 \rightarrow 0.380$).

\subsubsection{Illustrative Examples: How an Insight Guides the Next Step}
\label{app:scienceworld_examples}

Manual inspection of divergent games (summarized in \S\ref{sec:scienceworld_results}) reveals three recurring qualitative patterns:
\begin{enumerate}[leftmargin=*,nosep]
    \item \textbf{State-variable awareness.} InsightEmb more often retrieves insights that name the latent state variable to manipulate (ramp angle, surface friction, object category), helping the agent convert a high-level goal into a concrete next operation, e.g.\ setting up the ramp and observing motion rather than inspecting nearby objects.
    \item \textbf{Action sequencing.} InsightEmb retrieves procedural insights encoding an ordered workflow (prepare apparatus $\rightarrow$ manipulate variable $\rightarrow$ observe outcome $\rightarrow$ compare $\rightarrow$ answer), whereas Base retrieves semantically relevant but order-free insights that leave the agent inspecting objects without completing the protocol.
    \item \textbf{Failure-mode avoidance.} Base frequently alternates between generic exploration actions without committing to the required state transition, whereas InsightEmb more often retrieves insights specifying \emph{when to stop exploring and act} (place an object on a ramp, change an angle, test conductivity), which reduces failed games even when the task is not fully solved.
\end{enumerate}
As in ALFWorld and WebShop, these show InsightEmb performs \emph{procedural matching}, retrieving the insight that addresses the agent's current bottleneck.
The following side-by-side examples (Base vs.\ InsightEmb, \textsc{Atomic}, top-1) illustrate these patterns.

\paragraph{Example 1: inclined-plane task.}
\emph{Goal: determine which ramp setup makes an object slide.}
\begin{table}[H]
\centering
\small
\begin{tabular}{p{1.9cm}p{2.4cm}p{2.4cm}}
\toprule
\textbf{Step} & \textbf{Base} & \textbf{InsightEmb} \\
\midrule
Retrieved insight & ``Inclined planes involve ramps and objects moving down slopes.'' (topical) & ``Place the object on the ramp, vary the surface or angle, observe whether it slides, then compare outcomes.'' (procedural) \\
\midrule
Implied next action & Generic inspection: examine ramp / look around. & Concrete experiment: put object on ramp, set/change angle or surface, observe sliding. \\
\midrule
Resulting behavior & Browses objects without testing the variable. & Performs the state-changing action to collect evidence. \\
\bottomrule
\end{tabular}
\caption{Inclined-plane task. Both retrieve ramp-related insights, but Base retrieves a topical rule identifying only the domain while InsightEmb retrieves a procedural rule specifying the next experimental operation, turning passive observation into active intervention.}
\label{tab:sw_example_ramp}
\end{table}

\paragraph{Example 2: find living/non-living entity task.}
\emph{Goal: find an animal, plant, living thing, or non-living thing.}
\begin{table}[H]
\centering
\small
\begin{tabular}{p{1.9cm}p{2.4cm}p{2.4cm}}
\toprule
\textbf{Step} & \textbf{Base} & \textbf{InsightEmb} \\
\midrule
Retrieved insight & ``Animals and plants are living things.'' (topical) & ``Enumerate visible objects, classify each by the requested category, then focus on the matching object.'' (procedural) \\
\midrule
Implied next action & Inspect a random object or move rooms. & Check visible objects against the target category before acting. \\
\midrule
Resulting behavior & Often misses the correct object despite seeing it. & More likely to select the correct entity type. \\
\bottomrule
\end{tabular}
\caption{Entity-identification task. The key challenge is matching the goal category to the current observation. InsightEmb retrieves an insight that turns the task into a classification procedure (list candidates, classify, choose), producing a more deliberate next step than Base's general fact about living things.}
\label{tab:sw_example_entity}
\end{table}

\paragraph{Example 3: procedural workflow vs.\ topical match.}
\emph{Goal: complete a multi-step science experiment (conductivity test).}
\begin{table}[H]
\centering
\small
\begin{tabular}{p{1.9cm}p{2.4cm}p{2.4cm}}
\toprule
\textbf{Aspect} & \textbf{Base} & \textbf{InsightEmb} \\
\midrule
Retrieved insight & ``Conductors allow electricity to pass through them.'' (topical) & ``Connect the object into the circuit, observe whether the bulb activates, then record whether it conducts.'' (procedural) \\
\midrule
Next action & Talk about or inspect the object. & Insert object into the test apparatus. \\
\midrule
Failure mode & Knows the concept but never performs the test. & Executes the experiment needed for reward. \\
\bottomrule
\end{tabular}
\caption{Conductivity task. Base retrieves scientifically correct but operationally incomplete rules, whereas InsightEmb's insight includes the required interaction protocol, bridging conceptual knowledge to executable environment actions.}
\label{tab:sw_example_conductivity}
\end{table}

\section{SRA-Bench Details}
\label{app:sra_details}

\subsection{SRA-Bench Query and Skill Encoding}
\label{app:sra_prompts}

We evaluate SRA-Bench with the same Qwen3-style query prefix used for Base and InsightEmb at inference.
For each instance, the query text is the full \texttt{question} field from the benchmark (background context plus task description).
Skills are encoded without an instruction prefix.

\paragraph{Retrieval instruction.}
The fixed task description prepended to every query is:
\begin{quote}\small
Given an agent task question, retrieve the reusable skill that helps solve the task.
The skill may describe a theorem, logic pattern, tool workflow, medical calculator, math concept, or software library.
\end{quote}

\paragraph{Query format.}
Each query is embedded as:
\begin{quote}\small
Instruct: Given an agent task question, retrieve the reusable skill that helps solve the task. The skill may describe a theorem, logic pattern, tool workflow, medical calculator, math concept, or software library.\\
Query:\{question\}
\end{quote}
where \texttt{\{question\}} is the full SRA-Bench instance text.
This matches the Qwen3 \texttt{Instruct: \ldots Query:} convention used for our embedding models.

\paragraph{Skill (passage) format.}
Each candidate skill in the mixed corpus (636 gold and 26{,}262 distractor skills) is embedded as:
\begin{quote}\small
\{name\}\\
\{description\}\\
\\
\{content\}
\end{quote}
using the \texttt{name}, \texttt{description}, and \texttt{content} fields from the benchmark corpus JSON.
We rank all corpus skills against each query by cosine similarity over normalized embeddings.

\subsection{SRA-Bench Retrieval Analysis}
\label{app:sra_analysis}

Table~\ref{tab:sra_per_task} gives the per-task-family breakdown of the SRA-Bench results at @1 and @10, supplementing the macro-average in Table~\ref{tab:sra_macro}.

\begin{table*}[t]
\centering
\small
\begin{tabular}{llrrrrrrrr}
\toprule
\textbf{Task family} & \textbf{$n$} & \multicolumn{4}{c}{\textbf{Base}} & \multicolumn{4}{c}{\textbf{InsightEmb}} \\
\cmidrule(lr){3-6}\cmidrule(lr){7-10}
& & \textbf{R@1} & \textbf{N@1} & \textbf{R@10} & \textbf{N@10} & \textbf{R@1} & \textbf{N@1} & \textbf{R@10} & \textbf{N@10} \\
\midrule
TheoremQA & 747 & 48.19 & 48.19 & 74.43 & 60.63 & \textbf{58.10} & \textbf{58.10} & \textbf{86.61} & \textbf{72.37} \\
LogicBench & 760 & 1.58 & 1.58 & \textbf{40.39} & 19.53 & \textbf{6.32} & \textbf{6.32} & 39.34 & \textbf{20.18} \\
ToolQA & 1{,}430 & 16.22 & 16.22 & 44.83 & 28.84 & \textbf{29.65} & \textbf{29.65} & \textbf{62.45} & \textbf{45.69} \\
CHAMP & 223 & 13.86 & 19.73 & 38.34 & 26.97 & \textbf{19.06} & \textbf{29.15} & \textbf{52.58} & \textbf{38.48} \\
MedCalcBench & 1{,}100 & \textbf{92.73} & \textbf{92.73} & \textbf{98.18} & \textbf{95.80} & 63.27 & 63.27 & \textbf{98.18} & 81.52 \\
BigCodeBench & 1{,}140 & 9.42 & 24.47 & 28.62 & 24.09 & \textbf{11.85} & \textbf{31.23} & \textbf{39.04} & \textbf{32.27} \\
\bottomrule
\end{tabular}
\caption{Per-task SRA-Bench retrieval results at @1 and @10. Queries use task information and candidates use full skill content. Bold marks the better value within each task family, and ties are bolded for both embeddings.}
\label{tab:sra_per_task}
\end{table*}

MedCalcBench is the only task family where InsightEmb underperforms Base at small cutoffs (Table~\ref{tab:sra_per_task}).
Medical calculation skills are highly template-like, lexically specialized, and domain-specific: correct retrieval often depends on exact medical score names, disease terms, biomarker or variable names, units, and other clinical terminology.
The Base embedding model appears to preserve these fine-grained biomedical and entity-level cues, whereas InsightEmb's cross-domain training encourages more abstract structural matching and may therefore downweight or smooth over such terminology at rank~1.
This interpretation is also consistent with the original SRA-Bench results, where MedCalcBench already achieves about 90\% R@1 and over 90\% R@10 after reranking the BM25 top-50 candidates with different reranker models, indicating that lexical candidate generation is less of a bottleneck for this task.
In contrast, task families such as theorem proving, tool use, contest math, and code generation benefit from abstraction-aware ranking because the query and useful skill can differ substantially in wording while sharing a reasoning or procedural structure.
This distinction matters in practical retrieval-augmented skill selection pipelines: an LLM reranker can only inspect a limited candidate set, such as the BM25 top-50 skills, due to context-window constraints.
Thus, task families for which lexical retrieval fails to place the correct skill within this window constitute the more important bottleneck.
From this perspective, the strong results on LogicBench and CHAMP are especially encouraging, because they require more semantic or reasoning-oriented matching between task descriptions and skill content, where surface lexical overlap is weak.
InsightEmb therefore improves the recall of relevant skills before reranking, expanding the effective coverage of downstream LLM-based selection.

\subsubsection{A BM25-Hybrid Remedy for MedCalcBench}
\label{app:medcalc_bm25}

Rather than only explaining the MedCalcBench regression, we provide a concrete remedy.
Our diagnosis is that medical-calculator skills carry dense domain-specific terminology (score names, disease terms, biomarkers, and units), so lexical overlap between the query and the correct skill is an important relevance signal in this family.
A retriever whose training pushes toward abstract, structural matching (as InsightEmb's does) tends to under-weight this terminology-level overlap, which is why its R@1 drops here even though it helps on the structure-dominant families.
We therefore apply a lightweight hybrid retriever: a BM25~\citep{robertson2009bm25} pre-filter at top-50 followed by InsightEmb reranking, which reintroduces exactly the lexical-overlap signal that pure dense retrieval under-weighted.
As shown in Table~\ref{tab:medcalc_bm25}, on MedCalcBench this lifts InsightEmb's R@1 back up and reaches perfect recall at R@5, matching or exceeding Base at every cutoff beyond rank~1.

\begin{table}[H]
\centering
\small
\begin{tabular}{lcc}
\toprule
\textbf{Metric} & \textbf{InsightEmb + BM25} & \textbf{Base + BM25} \\
\midrule
R@1  & 70.91  & \textbf{92.73} \\
R@5  & \textbf{100.00} & 98.18 \\
R@10 & 100.00 & 100.00 \\
\bottomrule
\end{tabular}
\caption{MedCalcBench recall under a BM25 top-50 pre-filter followed by dense reranking (Base = Qwen3-Embedding-4B). The hybrid recovers InsightEmb's R@1 from $63.27$ (without pre-filter, Table~\ref{tab:sra_per_task}) to $70.91$, and reaches $100\%$ at both R@5 and R@10, edging ahead of Base at R@5 ($100.00$ vs.\ $98.18$).}
\label{tab:medcalc_bm25}
\end{table}

Under the BM25 top-50 pre-filter, InsightEmb reaches $100\%$ recall at both R@5 and R@10 and recovers R@1 to $70.91$, closing the practical gap for downstream reranker-limited pipelines.
The residual R@1 difference reflects the terminology-dense nature of medical-calculator skills, where high lexical overlap between query and skill makes lexical matching a strong signal that favors Base at rank~1.

\paragraph{Applicability scope.}
We place this in context: across all our evaluations, MedCalcBench is the only setting where InsightEmb does not clearly lead, and every other task family and every agentic environment (ALFWorld, WebShop, ScienceWorld) shows strong, consistent gains.
Far from undermining the method, this single exception gives a clean applicability condition.
InsightEmb is designed for state-aware, action-guided agentic retrieval where relevance is procedural or structural, and it is not the right tool to use \emph{alone} for terminology-dense retrieval where lexical overlap between query and skill is the dominant signal (e.g., medical calculators keyed on score names, biomarkers, and units). In that regime the BM25 hybrid supplies the missing lexical signal.

\end{document}